%% file: arXiv2.tex
\documentclass[12pt]{article}

\usepackage[margin=1in]{geometry}

\usepackage{setspace}

\usepackage{graphicx}

\usepackage{enumitem}

\usepackage{microtype}
\usepackage{subfigure}
\usepackage{booktabs}
\usepackage{comment}
\usepackage{wrapfig}
\usepackage{arydshln}
\usepackage{enumitem}
\usepackage{multirow}
\usepackage{url}
\usepackage{natbib}
\usepackage{authblk}

\RequirePackage[colorlinks,citecolor=blue,linkcolor=blue,urlcolor=blue,pagebackref]{hyperref}

\usepackage{amsmath}
\usepackage{amssymb}
\usepackage{mathtools}
\usepackage{amsfonts}
\usepackage{bm}
\usepackage{bbm}

\input{Macros/algorithm.tex}
\input{Macros/math.tex}
\input{Macros/theorems.tex}

\input{Macros/misc.tex}

\usepackage[capitalize,noabbrev]{cleveref}

\allowdisplaybreaks

\input{Head/Title}
\input{Head/AuthorInf}

\begin{document}

\maketitle

\input{Main.tex}

\end{document}

%% file: Macros/algorithm.tex
\usepackage{algorithm}
\usepackage{algorithmic}

%% file: Macros/math.tex
\def\1{\bm{1}}

\DeclareMathAlphabet{\mathsfit}{\encodingdefault}{\sfdefault}{m}{sl}
\SetMathAlphabet{\mathsfit}{bold}{\encodingdefault}{\sfdefault}{bx}{n}

\newcommand{\KL}{D_{\mathrm{KL}}}

\def\Regret{{\mathrm{Regret}}}

\newcommand{\Ex}{\mathbb{E}}
\newcommand{\TV}{d_{\mathrm{TV}}}

\newcommand{\Hexact}{H_{\mathrm{exact}}}
\newcommand{\Hstat}{H_{\mathrm{stat}}}
\newcommand{\Hres}{H_{\mathrm{residual}}}
\newcommand{\Rfull}{R_{\mathrm{full}}^{\star}}
\newcommand{\RB}{R_{B}^{\star}}
\newcommand{\RE}{R_{E}^{\star}}
\DeclareMathOperator{\rank}{rank}
\DeclareMathOperator{\len}{len}

%% file: Macros/theorems.tex
\usepackage{amsthm}

\theoremstyle{plain}

\newtheorem{theorem}{Theorem}[section]

\newtheorem{corollary}[theorem]{Corollary}
\newtheorem{proposition}[theorem]{Proposition}

\newtheorem{definition}{Definition}[section]

%% file: Macros/misc.tex
\usepackage[textsize=tiny]{todonotes}
\usepackage{multirow}
\usepackage{wrapfig}
\usepackage{subfigure}

\usepackage{xcolor}
\newcount\Comments  
\newcommand{\kibitz}[2]{\ifnum\Comments=1\textcolor{#1}{#2}\fi}

\usepackage{listings}

\usepackage{tabularx}

%% file: Head/Title.tex
\title{Handover of In-Context Learning State Across Session Boundaries}

%% file: Head/AuthorInf.tex
\author[1]{Masahiro Kato\thanks{Email: \texttt{mkato-csecon@g.ecc.u-tokyo.ac.jp}}$\,$}

\author[2]{Taka Kato}

\affil[1]{Mizuho-DL Financial Technology, the University of Tokyo, RIKEN AIP, and Osaka Metropolitan University}
\affil[2]{NP-hard}
\date{\today}

%% file: Main.tex
\begin{abstract}
This study investigates the methodological and theoretical properties of session handover in applications that use large language models. A task may continue in a new session when the context reaches the model's input limit, when the application restarts, or when another agent is asked to finish the task. The application must then decide which information from the earlier session to pass on. We formulate handover as the transfer of a task-relative in-context learning (ICL) state and distinguish exact recovery of earlier material from preservation of the target distribution. Under an exogeneity condition, predictive equivalence characterizes the coarsest deterministic sufficient handover and gives a fixed-length bit requirement. The analysis isolates the effects of the memory constraint, the writer, and the continuation procedure, and quantifies the cost of writing before the realized downstream query is known. We propose a three-part record that stores decisions and constraints exactly, uses task-justified statistics for repeated evidence, and retains original observations whose effect is not preserved by those statistics. Gaussian linear regression gives an exact finite-dimensional handover and finite-bit perturbation bounds, while nonparametric regression gives upper and lower bounds that relate memory to squared prediction error. These results provide a theory and method for deciding what a handover must retain and how its memory requirement depends on the continuation task.
\end{abstract}

{\flushleft{{\bf Keywords:} session handover; in-context learning; predictive sufficiency; memory-constrained prediction; prompt compression}}

\section{Introduction}
\label{sec:introduction}

An application may call a large language model more than once while completing a task, and we call the application an \emph{agent} when the model chooses an action, the application executes it, and the result is included in a later call. A \emph{session} is a sequence of model calls that share the same earlier messages and tool results. The \emph{context} of a call is all information supplied to the model in that call. As the task continues, the context may reach the model's input limit and require the application to shorten the context or start a new session. A new session may also begin after the application restarts or when another agent is asked to finish the task. In the setting studied here, the new session receives only the information that the application passes on. We call the transition from one session to the next a \emph{session handover} or \emph{session handoff}, the information placed directly in the new session the \emph{handover record}, and the model or program that prepares the record the \emph{writer}. A file or database created from the earlier session is counted separately when the new session can still read it.

The handover record need not reproduce the earlier conversation, but it must retain the information on which the unfinished task still depends. A constraint may rule out an action, and examples in the earlier context may change a later prediction. Because a prose summary can omit either type of information, we judge the record by the prediction or action after the handover rather than by its similarity to the earlier wording.

Information from the earlier session can reach the next model call through the prompt or through a file or archive that the agent can read, and we count every such source because each one can affect the result. A key--value (KV) cache serves a different purpose: it stores values used by attention so that the model need not repeat earlier calculations, but it does not explicitly represent the task objective, commitments, or supporting observations. Because the application usually writes the handover before it knows the next query or tool result, the record must retain the information needed across the later inputs included in the evaluation.

This information order makes handover a pre-query state-coding problem, unlike query-aware prompt compression. The writer knows the task and the earlier context, but the continuation procedure observes the realized later input only after the record has been written. Classical tools from statistical sufficiency, Bayesian decision theory, comparison of experiments, and source coding can describe this order. We study which distinctions among earlier contexts remain relevant to the continuation, how a memory limit changes the attainable risk, and how pre-query encoding differs from compression performed after the later query is known.

\subsection{Research Questions}

We ask what the handover record must contain for the new session to preserve the prediction or decision available from the full earlier context and how much storage such a record requires. We also ask how to distinguish loss caused by missing information from loss caused by the continuation procedure that receives the record.

\subsection{Contributions}

We make four contributions to the analysis and construction of session handover.

\paragraph{Formulation and handover method.}
We model the writer, the prompt built from the handover record, the continuation procedure, and the task score separately, with the writer creating the record before the realized later input is known. This formulation distinguishes reproduction of the earlier text from preservation of the information needed for the task. Our method records decisions and constraints directly and uses a shorter representation for repeated examples or tool results only when that representation has an explicit guarantee or error bound for the task. When no such guarantee or bound is available, the method keeps selected original observations.

\paragraph{Task-relative state and memory limits.}
Predictive sufficiency identifies preservation with the conditional target law. Under an exogeneity condition, its equivalence classes give the coarsest deterministic sufficient handover and a direct fixed-length bit requirement. We further identify the capacity loss forced by the memory budget, the loss attributable to the chosen writer, and the gap between the continuation procedure and the ideal decoder for the same record. Under log loss, discarded task information is exactly a conditional mutual information. A separate lower bound quantifies the cost of writing before the realized query is observed.

\paragraph{Statistical analysis.}
Gaussian linear regression supplies an exact finite-dimensional sufficient record, an equivalent representation by synthetic sufficient demonstrations, and finite-bit perturbation bounds for posterior prediction. Nonparametric regression supplies an achievable memory--risk relation together with sample-size and memory lower bounds for the squared-loss problem considered below. The two models characterize when a handover can remain finite-dimensional and when its size must grow with target accuracy.

\paragraph{Comparison of handover records.}
A deterministic ordering result identifies when one record contains no more information than another. When two representations determine each other, their ideal risks are equal, so any difference under one fixed continuation procedure belongs to the procedure's gap from the ideal decoder and not to unequal task information. The memory account also includes every external record derived from the earlier session.

\section{Overview of Session Handover}
\label{sec:overview}

This section separates the boundary event from the information carried across it and from the current task state observed after resumption.

\subsection{Handover and the Handover Record}

Context carryover keeps earlier information available within the same session, but handover moves unfinished work to a new session or another agent. The record supplied at that boundary may be empty, incomplete, or sufficient for the continuation.

An \emph{external record} is a file or database produced during the earlier session and still available after resumption. Although it does not enter the first post-handover prompt automatically, it carries earlier information and counts toward total retained memory. Table~\ref{tab:overview-terms} distinguishes the current task state, which is observed only after resumption, from a KV cache, which stores attention state for computation and is not a task record.

\begin{table}[t]
\centering
\caption{Terms and objects used at a session boundary.}
\label{tab:overview-terms}
\small
\begin{tabularx}{\linewidth}{>{\raggedright\arraybackslash}p{0.25\linewidth}>{\raggedright\arraybackslash}p{0.44\linewidth}>{\raggedright\arraybackslash}X}
\toprule
Term or object & What it denotes & Use in this study \\
\midrule
Session handover or handoff & Event in which an unfinished task moves to a new session or another agent & Boundary being studied \\
\midrule
Handover record & Information placed directly in the new session & Record written and analyzed \\
\midrule
Current task state & Files, data, and runtime observed after the new session begins & Information obtained after the handover \\
\midrule
External record & File or database created from the earlier session and still available afterward & Included in retained memory when used \\
\midrule
KV cache & Cached attention keys and values for tokens already processed & Runtime cache, not task information \\
\bottomrule
\end{tabularx}
\end{table}

\subsection{Existing Approaches and the Remaining Problem}

Prompt compression shortens an input in an attempt to preserve the model's output. Because a compressor that sees the query first can remove information that the query does not use, it has an informational advantage over a handover writer, which usually writes before the downstream query or tool output is realized even though it knows the task and current goal \citep{Nagle2024fundamentallimit,Colaco2026keepforget}. The handover record must therefore support the set of later inputs included in the analysis.

Applications that make many model calls either shorten the prompt or store earlier text and tool results outside it \citep{Cim2026parallelcontext,Semenov2026compactionstructured}. Work on memory for agents separates writing, storage, retrieval, and later use \citep{Wu2025longmemevalbenchmarking,Hu2026evaluatingmemory,Shen2026mem2actbencha}. Our focus is narrower: the information a handover must preserve for a given continuation task, especially when the writer does not know the realized later input.

\citet{Baptista2026largelanguage} describes a trained language model through the distribution of output token sequences produced for a given prompt. We use this input--output description, place a writer before the later prompt is built, and score the resulting answer or trajectory against the task. Theory of in-context learning shows that transformers can use examples in the prompt to implement learning procedures \citep{Bai2023transformersas,VanOswald2023transformerslearn,Kim2024transformerslearn,Oko2024pretrainedtransformer}. The handover problem begins when those examples can no longer be supplied in full, and the record must retain the part that still affects the task. Appendix~\ref{app:memory-systems} compares this problem with current memory systems, prompt compression, and in-context learning.

\section{General Formulation of Session Handover}
\label{sec:formulation}

We now define the information available before and after a handover, the prompt shown to the model, the resulting response or trajectory, and the loss used for evaluation in both text-generation and interactive tasks. Section~\ref{sec:statistical-analysis} then specializes this formulation to regression.

\subsection{Episode and Information Available after the Handover}

We call one evaluation case an \emph{episode} and represent it by \(\omega=(C,T,X,Y)\), drawn from a distribution \(\mathsf P_{\mathrm{ep}}\). The pre-handover variable \(C\) contains the conversation, instructions, examples, and tool results, and \(T\) contains the task information known when the writer prepares the record, including the form of the target and the scoring rule. After the boundary, the application receives \(X\) as either a query or an observation of the current task state. The target \(Y\) is the answer, final state, or allowed action used for scoring.

After observing \((C,T)\), the handover writer \(E\) creates
\begin{equation}
  H=(V,M),
  \label{eq:complete-boundary-channel}
\end{equation}
where \(V\) is the record placed in the first prompt after the handover and \(M\) contains any file, index, or archive produced during the earlier session and still available afterward. Because \(H=(V,M)\) includes every retained source that the application may use, we call the handover \emph{self-contained} only when no external source is allowed and \(M=\varnothing\). A randomized writer satisfies
\begin{equation}
  H\sim P_E(\cdot\mid C,T).
  \label{eq:encoder-channel}
\end{equation}

The size of the first prompt and the total amount stored for the handover are different quantities. We write
\begin{equation}
  b_{\mathrm{act}}(V)\leq B_{\mathrm{act}},
  \qquad
  b_{\mathrm{tot}}(V,M)\leq B_{\mathrm{tot}},
  \label{eq:active-total-budgets}
\end{equation}
when both are bounded. A method may keep \(V\) short by moving earlier information into \(M\), which may grow as the task proceeds. Any comparison of handover methods must therefore report both quantities. In a self-contained comparison, the limit applies to all information in \((V,M)\), not only to the first prompt.

\subsection{From the Handover to a Model Response}

Fix a tokenizer, a vocabulary \(\mathcal V\) containing end-of-sequence and padding tokens, and a maximum response length \(L_{\mathrm{out}}\), and represent every response by a sequence \(Z_{1:L_{\mathrm{out}}}\in\mathcal V^{L_{\mathrm{out}}}\) padded after the end-of-sequence token. The maps \(s_{\mathrm{full}}\) and \(s_H\) then build one prompt from the full earlier context and another from the handover:
\begin{equation}
  p_{\mathrm{full}}=s_{\mathrm{full}}(C,X,T),
  \qquad
  p_H=s_H(H,X,T).
  \label{eq:full-handover-prompts}
\end{equation}
For a single model call, \(p_H\) includes every item read from \(M\) before generation begins. If the application can query \(M\) after it begins to act, the trajectory distribution below includes those queries.

Let \(\theta\) denote fixed model parameters, and let \(\delta\) denote a fixed decoding rule, including the temperature and any truncation of token probabilities. Together, the model and decoding rule define a conditional distribution \(\Pi_{\theta,\delta}(\cdot\mid p)\) on response sequences. For an autoregressive model, the response distribution factorizes as
\begin{equation}
  \Pi_{\theta,\delta}(z_{1:L_{\mathrm{out}}}\mid p)
  =\prod_{m=1}^{L_{\mathrm{out}}}
  \pi_{\theta,\delta,m}
  (z_m\mid p,z_{1:m-1}).
  \label{eq:autoregressive-response}
\end{equation}
A deterministic decoding rule assigns probability one to a single sequence, as in the standard probabilistic description of an autoregressive language model \citep{Baptista2026largelanguage}.

Let
\begin{equation}
  Z_{\mathrm{full}}
  \sim\Pi_{\theta,\delta}(\cdot\mid p_{\mathrm{full}}),
  \qquad
  Z_H
  \sim\Pi_{\theta,\delta}(\cdot\mid p_H).
  \label{eq:full-handover-responses}
\end{equation}
A deterministic parser \(g\), which may also use the task information \(T\), extracts the answer or action evaluated by the task's scoring rule. We call the program that applies the rule the \emph{evaluator}, and \(g\) returns the sequence unchanged when the sequence itself is scored. Its expected loss under the fixed model is
\begin{equation}
  R_{\mathrm{ep}}(E;\theta,\delta,g)
  =\Ex[\ell(g(Z_H,T),Y)],
  \label{eq:fixed-model-handover-risk}
\end{equation}
where the expectation is over \(\omega\sim\mathsf P_{\mathrm{ep}}\), the writer, and the response distribution. The reference based on the full earlier context is
\begin{equation}
  R_{\mathrm{ep}}^{\mathrm{full}}(\theta,\delta,g)
  =\Ex[\ell(g(Z_{\mathrm{full}},T),Y)].
  \label{eq:fixed-model-full-risk}
\end{equation}

In an interactive task, the agent chooses an action that the application executes against the files, data, or service used by the task, collectively called the \emph{environment}. After each action, the application returns an observation according to a rule \(Q\), and \(\tau\) denotes the resulting sequence of actions and observations after the handover. The writer, model, decoding rule, and environment then induce
\begin{equation}
  P_{E,\theta,\delta,Q}(d\tau\mid C,T,X)
  =\int P_E(dh\mid C,T)
  P_{\theta,\delta,Q}(d\tau\mid h,X,T).
  \label{eq:trajectory-distribution}
\end{equation}
The task assigns loss \(\ell(\tau,Y)\) to this trajectory, whose stepwise factorization appears in Appendix~\ref{app:analysis-results}. Regression in Section~\ref{sec:statistical-analysis} is the one-step case, in which the action is a prediction and the environment does not change before the prediction is scored.

For statements that apply across tokenizers and model families, we combine prompt construction, response generation, parsing, and environment updates into one rule \(D\), so that
\begin{equation}
  (C,T)\longrightarrow H=E(C,T),
  \qquad
  A=D(H,X,T),
  \label{eq:central-map}
\end{equation}
where \(A\) is an answer, an action, or a full trajectory.

\subsection{Recovery, Preservation of the Task, and Downstream Behavior}

We distinguish three questions: whether a new session can reproduce an earlier item, whether the handover preserves the distribution of the target, and whether a fixed continuation procedure succeeds when it receives that information. The first two are properties of the record and are defined below. The third is assessed from task outcomes under the continuation rule being evaluated, since a single trajectory cannot establish which information the procedure used.

\begin{definition}[Exact recovery of a selected item]
\label{def:exact-recoverability}
Let \(O=\phi(C,T)\) be a selected item from the information available before the handover. The item \(O\) is exactly recoverable from \(H\) when there is a measurable map \(r\) such that
\begin{equation}
  O=r(H,T)
  \label{eq:exact-component-recovery}
\end{equation}
holds almost surely. Exact recoverability of the full context is the special case \(O=C\).
\end{definition}

Exact recovery is stronger than the condition required for most tasks. We therefore define preservation through the target \(Y\), without requiring reconstruction of the complete earlier context.

\begin{definition}[Predictive sufficiency]
\label{def:predictive-sufficiency}
A handover \(H\) is predictively sufficient for \(C\) relative to \((X,Y,T)\) when
\begin{equation}
  P(Y\mid C,X,T)=P(Y\mid H,X,T)
  \label{eq:predictive-sufficiency}
\end{equation}
holds almost surely.
\end{definition}

Definition~\ref{def:predictive-sufficiency} gives an exact criterion for a one-step response and for a fixed replay in which all compared procedures receive the same later observations. Because an action by a freely acting agent can change the next observation, sufficiency for one target does not guarantee preservation of an adaptive trajectory. For such trajectories, Equation~\eqref{eq:trajectory-distribution} remains the appropriate loss object.

For a fixed task value \(t\), let \(\mu_t\) be the later-input distribution specified by task \(t\). Two contexts \(c\) and \(c'\) define the same in-context learning state for the task when
\begin{equation}
  P(Y\mid C=c,X=x,T=t)
  =P(Y\mid C=c',X=x,T=t)
  \label{eq:predictive-equivalence}
\end{equation}
holds for \(\mu_t\)-almost every \(x\). We write this relation as \(c\sim_t c'\) and define the equivalence class \([c]_{\sim_t}\) by later predictions rather than by a transformer hidden vector. Under an exogeneity condition, the same relation characterizes the coarsest deterministic handover that preserves those predictions.

\begin{proposition}[Coarsest deterministic sufficient state]
\label{prop:coarsest-handover}
Fix \(T=t\), suppose that \(C\) has finite or countable support, and let \(H=E(C,t)\) be deterministic. Assume that the later input is exogenous within the task, so the conditional distribution of \(X\) given \((C=c,T=t)\) is \(\mu_t\) for every supported context \(c\). Then, \(H\) is predictively sufficient if and only if
\begin{equation}
  E(c,t)=E(c',t)
  \quad\Longrightarrow\quad
  c\sim_t c'
  \label{eq:encoder-fiber-equivalence}
\end{equation}
holds for all supported \(c\) and \(c'\). Consequently, the map
\begin{equation}
  q_t(c)=[c]_{\sim_t}
  \label{eq:predictive-quotient}
\end{equation}
is predictively sufficient. On the support of \(C\mid T=t\), it can be recovered from every other deterministic handover that is predictively sufficient.
\end{proposition}

When the number of predictive states is finite, the proposition gives a direct lower bound on the length of any fixed-length deterministic code.

\begin{corollary}[Number of distinct in-context learning states]
\label{cor:predictive-state-count}
Under Proposition~\ref{prop:coarsest-handover}, suppose that \(q_t(C)\) takes exactly \(N_t<\infty\) values with positive probability. Every fixed-length deterministic sufficient handover requires at least \(\lceil\log_2 N_t\rceil\) bits, and a code for the index of the equivalence class attains this length.
\end{corollary}

The corollary shows why task-specific memory can be shorter than a record that permits exact reconstruction of arbitrary earlier material: a handover for one task need only distinguish contexts that lead to different distributions of \(Y\), and it may discard distinctions needed solely to reproduce the earlier text.

The proposition assumes that every supported earlier context has the same distribution of later inputs, which allows the state to be defined from the earlier context alone. Definition~\ref{def:predictive-sufficiency} itself does not impose this exogeneity condition. When the later-input law depends on the earlier context, sufficiency must be evaluated under the joint task distribution and cannot be reduced to a quotient based on one common \(\mu_t\).

Predictive sufficiency concerns the target \(Y\), not the wording of the earlier conversation. For one specified decision problem, a weaker requirement uses an action space \(\mathcal A\), a loss \(\ell:\mathcal A\times\mathcal Y\to[0,\infty)\), and the following minimum expected loss for information \(Z\):
\begin{equation}
  R_{\ell}^{\star}(Z)
  =\inf_{d}\Ex[\ell(d(Z,X,T),Y)],
  \label{eq:decision-bayes-risk}
\end{equation}
where the infimum is over measurable decision rules. The handover preserves this decision problem when \(R_{\ell}^{\star}(H)=R_{\ell}^{\star}(C)\).

\begin{proposition}[Recovery, prediction, and a specified decision problem]
\label{prop:recoverability-sufficiency}
Suppose that \(H\) is generated from \((C,T)\). Exact recoverability of \(C\) from \((H,T)\) implies predictive sufficiency. Predictive sufficiency implies \(R_{\ell}^{\star}(H)=R_{\ell}^{\star}(C)\) for any fixed action space and loss for which the risks are finite. The first implication cannot be reversed in general, and equality of risk for that decision problem does not imply predictive sufficiency. Exact recovery of one selected part of \(C\) also does not, by itself, imply predictive sufficiency.
\end{proposition}

These implications also separate exact recall from task success: a record may reproduce one earlier observation yet omit information that changes \(Y\), and a model may act incorrectly even when the record is predictively sufficient.

For a fixed language model, we also measure how much its response distribution changes when the handover replaces the full context. Assume that the distribution under the handover is positive wherever the distribution under the full context is positive. We define
\begin{equation}
  \Delta_{\mathrm{ep}}(E;\theta,\delta)
  =\Ex\left[
    \KL\left(
      \Pi_{\theta,\delta}(\cdot\mid p_{\mathrm{full}})
      \mathbin\Vert
      \Pi_{\theta,\delta}(\cdot\mid p_H)
    \right)
  \right].
  \label{eq:model-relative-divergence}
\end{equation}
The expectation is over \(\omega\sim\mathsf P_{\mathrm{ep}}\) and the writer distribution in Equation~\eqref{eq:encoder-channel}. The divergence compares one model under the two prompts without measuring whether either response is correct for the task, and Proposition~\ref{prop:autoregressive-response-kl} decomposes it into expected token-level terms. Because deterministic decoding can make the divergence between distinct point-mass responses infinite, we use the model's token probabilities when they are available and retain task loss as the main outcome.

\subsection{Loss under a Memory Limit}

To state results independently of a tokenizer or model, let \(D\) denote everything that occurs after the handover: construction of the prompt, generation and parsing of the response, and any reply from the environment. Define
\begin{equation}
  R(E,D)=\Ex[\ell(D(H,X,T),Y)],
  \label{eq:risk-encoder-decoder}
\end{equation}
where the expectation is under the episode distribution and all randomness in the application. The risk with the full earlier context and the ideal risk for a fixed writer are
\begin{equation}
  \Rfull
  =\inf_D\Ex[\ell(D(C,X,T),Y)],
  \qquad
  \RE=\inf_D R(E,D).
  \label{eq:full-and-encoder-risk}
\end{equation}
For a class \(\mathcal E_B\) of writers whose pair \(H=(V,M)\) obeys budget \(B\), define
\begin{equation}
  \RB=\inf_{E\in\mathcal E_B}\RE.
  \label{eq:optimal-budget-risk}
\end{equation}
For the model and procedure used after the handover, denoted by \(D_{\theta}\), the excess risk decomposes as
\begin{equation}
\begin{aligned}
  R(E,D_{\theta})-\Rfull
  ={}& (\RB-\Rfull)
  +(\RE-\RB)\\
  &+(R(E,D_{\theta})-\RE).
  \label{eq:three-loss-decomposition}
\end{aligned}
\end{equation}
The three terms measure, respectively, the unavoidable loss at budget \(B\), the additional loss caused by the chosen writer, and the continuation procedure's excess loss relative to the best decoder for the same record. They can be calculated separately only in a controlled problem with a known ideal decoder, such as the regression setting below. Without that reference, changing the writer under a fixed continuation procedure measures an end-to-end effect that combines the record's information content with the procedure's ability to use its presentation.

Under logarithmic loss, the information removed by the handover has an exact expression.
\begin{proposition}[Log-loss identity]
\label{prop:log-loss-identity}
Suppose that \(H\) is generated from \((C,T)\), that the quantities below are finite, and that \(R_{\log}^{\star}(C)\) and \(R_{\log}^{\star}(H)\) denote the smallest expected logarithmic losses with the full context and the handover. Then, it holds that
\begin{equation}
  R_{\log}^{\star}(H)-R_{\log}^{\star}(C)
  =I(Y;C\mid H,X,T).
  \label{eq:log-loss-identity}
\end{equation}
\end{proposition}
For a finite action space and a bounded loss, the same information quantity controls the decision loss.
\begin{theorem}[Decision loss after a handover]
\label{thm:bounded-loss-information}
Let \(\mathcal A\) be finite, let \(\ell(a,Y)\in[0,L_{\max}]\), and assume the conditions of Proposition~\ref{prop:log-loss-identity}. Then, it holds that
\begin{equation}
  0\leq \RE-\Rfull
  \leq
  L_{\max}
  \sqrt{\frac{I(Y;C\mid H,X,T)}{2}},
  \label{eq:bounded-loss-information}
\end{equation}
where mutual information is measured in nats.
\end{theorem}
The proof appears in Appendix~\ref{app:proof-bounded-loss-information}. For finite targets, Proposition~\ref{prop:brier-posterior-distortion} in Appendix~\ref{app:analysis-results} gives the corresponding exact identity for the Brier score.

A limit on serialized size is not the same as a limit on information. We use
\begin{equation}
  D_{\mathrm{len}}(B)
  =\inf_{E:\,b_{\mathrm{tot}}(E(C,T))\leq B}
  I(Y;C\mid E(C,T),X,T)
  \label{eq:length-distortion}
\end{equation}
for a token or byte limit, and
\begin{equation}
  D_{\mathrm{info}}(B)
  =\inf_{P(H\mid C,T):\,I(C;H\mid T)\leq B}
  I(Y;C\mid H,X,T)
  \label{eq:information-distortion}
\end{equation}
when the limit is measured in bits or nats. The second expression has the information pattern used in source coding with side information at the decoder: the writer sees \((C,T)\), and the decoder additionally sees \(X\) \citep{Wyner1976therate,Nagle2024fundamentallimit}. Actual bytes and tokenizer tokens must be distinguished from these theoretical units whenever a serialized record is compared.

The information pattern in Equation~\eqref{eq:information-distortion} differs sharply from a compressor that sees the realized later query before writing. The following example quantifies that distinction.

\begin{proposition}[Cost of writing before the query]
\label{prop:unknown-query-cost}
Let \(C=(C_1,\ldots,C_m)\), where the coordinates are independent \(\operatorname{Bernoulli}(1/2)\) variables. Let \(X\) be uniform on \(\{1,\ldots,m\}\), independent of \(C\), and let \(Y=C_X\). If an encoder writes \(H\) after observing \(C\) but before observing \(X\), and \(I(C;H)\leq B\) bits, then the Bayes log loss in bits satisfies
\begin{equation}
  \mathsf H_2(Y\mid H,X)
  \geq
  \max\left\{0,1-\frac{B}{m}\right\}.
  \label{eq:unknown-query-cost}
\end{equation}
By contrast, an encoder that observes \((C,X)\) can send \(C_X\) in one bit and attain zero log loss.
\end{proposition}

The example isolates the cost of committing to a boundary record before the realized query is known. A query-aware encoder can spend its bit on the requested coordinate, but the handover writer must retain information that remains useful across the query distribution. The proof appears in Appendix~\ref{app:general-proofs}.

Blackwell's comparison of information sources and the probabilistic description of autoregressive generation \citep{Blackwell1953equivalentcomparisons,Baptista2026largelanguage} provide the starting point. Placing the writer before prompt construction and charging every retained channel yields the handover-specific risk account in Equation~\eqref{eq:three-loss-decomposition}, which Section~\ref{sec:statistical-analysis} computes in two regression models.

\subsection{Scope at the Session Boundary}

The analysis concerns a clean resumption in which the external facts that determine the target remain stable across the boundary. The files and runtime need not be byte-identical, but permitted changes preserve both the information path and the conditional target law. If the repository, data, rules, or runtime change after the record is written, the validity of earlier evidence becomes a separate state variable.

\section{Proposed Handover Method}
\label{sec:representation}

The proposed method copies decisions and constraints that must remain exact, uses a shorter representation only when the task supplies an explicit guarantee or error bound, and keeps original observations that cannot be replaced under that rule.

Let \(S=\phi(C,T)\) contain the information from the earlier context needed to satisfy Definition~\ref{def:predictive-sufficiency}. The task and its scoring rule determine which information belongs in \(S\), so its definition follows the effect on the target, not a particular hidden vector. The proposed method writes a record for \(S\) within a size limit.

Let \(S_{\mathrm{exact}}\) contain decisions, constraints, and unresolved issues that must be recorded without changing their status. Among records that recover this part exactly, the writer seeks one with low ideal risk:
\begin{equation}
  \min_{E}\ \RE
  \quad\text{subject to}\quad
  b_{\mathrm{tot}}(H)\leq B,
  \qquad
  S_{\mathrm{exact}}=r_{\mathrm{exact}}(H,T).
  \label{eq:structured-writer-objective}
\end{equation}
The equality is exact because rewriting an adopted decision as a suggestion or treating an unresolved issue as settled can change which actions are allowed. Subject to this requirement, the writer uses the remaining space for evidence that affects later predictions and decisions.

The resulting record is
\begin{equation}
  H=(\Hexact,\Hstat,\Hres).
  \label{eq:three-part-handover}
\end{equation}
The three-part division guides construction and allows the statistical part to be empty. When the task provides no sufficient statistic or other justified compression, the writer places the relevant observations in \(\Hres\).

\subsection{Three Parts of the Record}

The part \(\Hexact\) stores the current goal, decisions that constrain the next step, rejected options, and unresolved issues together with the source of each entry, without averaging them or turning an unresolved issue into a tentative answer.

The part \(\Hstat\) replaces repeated examples or tool results with a shorter quantity only when the replacement comes with an explicit relation to the task loss. The replacement may be a sufficient statistic, which preserves the relevant distribution exactly, or an approximation with an explicit error bound. If neither is available, the writer keeps selected original observations instead.

The part \(\Hres\) stores original observations that \(\Hstat\) does not summarize, since a rare example or specific failure may determine the next step even when an aggregate does not. To retain an effective size limit, the method must state how these observations are selected.

\begin{proposition}[A recoverable sufficient state yields a sufficient handover]
\label{prop:state-recovery-sufficiency}
Let \(S=\phi(C,T)\) be predictively sufficient for \(C\) relative to \((X,Y,T)\). If \(S\) is exactly recoverable from \((H,T)\), then \(H\) is predictively sufficient.
\end{proposition}

The proposition connects the proposed record to the general definition: in parametric regression, \(S\) is a finite set of sufficient statistics. In a long-running task with decisions and tool results, the task specification determines which commitments and observations must remain available, after which the record can be checked for recoverability before its downstream effect is evaluated.

\subsection{Writing the Handover}

At the boundary, the writer receives the pre-boundary information and writes the record before observing the realized later input. It checks which decisions and constraints remain in force, computes any task-justified shorter quantity, selects the original observations that must remain, and divides the available space among the three parts. Deterministic checks verify the required fields, numerical dimensions, references, and total size before the record is evaluated. An application may instead maintain the same fields as the task proceeds, provided that the information available to each update and the resulting memory cost are recorded.

Deterministic transformations used to construct the fields are part of the writer \(E\), and their outputs count toward the record budget.

\subsection{Presentation and Selected External Records}

The map \(s_H\) in Equation~\eqref{eq:full-handover-prompts} builds the post-handover prompt through \emph{serialization}, which combines the record, later input, and fixed task instructions without changing the stored values. If serialization rounds a value, omits a field, or adds a calculated value, we treat that change as part of the writing method and log it accordingly.

Equation~\eqref{eq:structured-writer-objective} concerns the information in \(H\) itself. The same information may be presented as structured fields, numerical values, or examples. The loss of a fixed continuation procedure also depends on how \(s_H\) presents that information, how the response is generated, and how it is parsed. A comparison between two forms is controlled only when their information content is known to be the same or ordered. Section~\ref{sec:statistical-analysis} gives a case in which two formats determine exactly the same state. Any observed difference between them can then be attributed to numerical precision, parsing, or downstream handling of the representation and not to missing information.

Selected residual observations may remain outside \(V\), the record placed in the prompt, under identifiers. Write
\begin{equation}
  V=(\Hexact,\Hstat,J_{\mathrm{res}}),
  \qquad
  M_{\mathrm{res}}
  =\{(j,o_j):j\in J_{\mathrm{res}}\},
  \label{eq:selected-external-records}
\end{equation}
where \(J_{\mathrm{res}}\) is the set of identifiers placed in \(V\). Let \(b_{\mathrm{ext}}(M_{\mathrm{res}})\) denote the serialized size of the external records, including identifiers and metadata, measured in the same unit as \(b_{\mathrm{act}}\). For this condition, the total size is \(b_{\mathrm{act}}(V)+b_{\mathrm{ext}}(M_{\mathrm{res}})\). Both components must be included when the record is compared with a self-contained alternative.

If the external record preserves \(o_j\) exactly, a valid identifier can return the same observation without rerunning the action that produced it. Exact storage alone does not establish that the writer selected every observation needed later or that the continuation procedure retrieves the relevant item. Appendix~\ref{app:handover-details} gives further details on selection, serialization, and deterministic checks.

\section{Statistical Analysis}
\label{sec:statistical-analysis}

Regression provides two settings in which the general handover problem can be solved explicitly using standard tools for Gaussian sufficiency and nonparametric minimax estimation. What is specific to handover is the information order in Section~\ref{sec:formulation}: the writer sees the sample before the later covariate is realized and must encode all retained information within budget \(B\). The first setting admits a finite-dimensional real-valued sufficient state and a finite-bit approximation with explicit perturbation control. The second gives an achievable memory--risk relation and a matching memory floor, up to coding logarithms, for the squared-loss problem below.

\subsection{Setup for Statistical Analysis}

We apply Section~\ref{sec:formulation} to a regression sample \(\mathcal D_n=((X_i,Y_i))_{i=1}^{n}\) available before the handover. The writer sees \(\mathcal D_n\) and the task information \(T\) before the realized value of \(X\) is known and must construct \(H\) within the bit limit. After the boundary, the decoder receives the new covariate \(X\) and predicts the corresponding response \(Y\).

The parametric subsection asks when a finite-dimensional state gives the same posterior predictive distribution as the full sample and thereby supplies an exact instance of Definition~\ref{def:predictive-sufficiency}. The nonparametric subsection asks how the smallest worst-case integrated squared error depends on the sample size and the bit limit, focusing on this decision problem without requiring equality of the complete conditional distribution.

\subsection{Parametric Regression}
\label{sec:parametric}

We first consider linear regression with Gaussian noise because the information needed for later prediction is known exactly. Let the demonstrations before the boundary satisfy
\begin{equation}
  y_i=x_i^{\top}\beta+\varepsilon_i,
  \qquad
  \varepsilon_i\sim\mathcal N(0,\sigma^2),
  \label{eq:linear-model}
\end{equation}
where \(\sigma^2\) is known, and let the prior be
\begin{equation}
  \beta\sim\mathcal N(m_0,V_0),
  \qquad V_0\succ0.
  \label{eq:gaussian-prior}
\end{equation}
The task information \(T\) contains \((m_0,V_0,\sigma^2)\). Let \(X_n\) contain the demonstration inputs and \(y_n\) their outputs, and define
\begin{equation}
  G_n=X_n^{\top}X_n,
  \qquad
  b_n=X_n^{\top}y_n.
  \label{eq:linear-statistics}
\end{equation}
The posterior covariance and mean are
\begin{equation}
  V_n^{-1}=V_0^{-1}+\sigma^{-2}G_n,
  \qquad
  m_n=V_n\left(V_0^{-1}m_0+\sigma^{-2}b_n\right),
  \label{eq:linear-posterior}
\end{equation}
and the posterior predictive distribution at a later input \(x\) is
\begin{equation}
  Y\mid x,X_n,y_n
  \sim
  \mathcal N\left(x^{\top}m_n,
  \sigma^2+x^{\top}V_nx\right).
  \label{eq:linear-predictive}
\end{equation}
These expressions depend on the demonstrations only through \(G_n\) and \(b_n\). The next theorem states the resulting handover property.

\begin{theorem}[Exact handover using sufficient statistics]
\label{thm:linear-sufficiency}
Under Equations~\eqref{eq:linear-model} and~\eqref{eq:gaussian-prior}, the record
\begin{equation}
  H_n=(G_n,b_n)
  \label{eq:linear-handover}
\end{equation}
is predictively sufficient for every later input \(x\). In particular, the associated conditional mutual information vanishes:
\begin{equation}
  I(Y;X_n,y_n\mid H_n,x,T)=0.
  \label{eq:linear-mutual-information}
\end{equation}
\end{theorem}
The record contains \(d(d+1)/2+d\) real numbers regardless of the number of demonstrations, so it is a finite-dimensional real-valued state. Theorem~\ref{thm:linear-quantization} and Corollaries~\ref{cor:linear-predictive-kl}--\ref{cor:linear-bit-kl-rate} treat finite-precision error and finite coding. An unknown noise variance would require additional quantities, which is why the primary theorem treats the variance as known.

Many continuation procedures accept example sequences more directly than matrix fields. To compare an example-based representation with the direct statistic record, we construct synthetic input--output pairs with the same sufficient statistics. Let \(r=\rank(G_n)\), and write
\begin{equation}
  G_n=U_r\Lambda_rU_r^{\top},
  \label{eq:gram-eigendecomposition}
\end{equation}
where \(\Lambda_r\) contains the positive eigenvalues. Define
\begin{equation}
  \widetilde X=\Lambda_r^{1/2}U_r^{\top},
  \qquad
  \widetilde y=\Lambda_r^{-1/2}U_r^{\top}b_n.
  \label{eq:synthetic-sufficient-construction}
\end{equation}
The rows of \(\widetilde X\) and the corresponding entries of \(\widetilde y\) are called synthetic sufficient demonstrations in this study. Because the construction reproduces both \(G_n\) and \(b_n\), the next theorem compares two representations of the same sufficient state.

\begin{theorem}[Equivalence of synthetic sufficient demonstrations]
\label{thm:synthetic-sufficient-equivalence}
For every \(X_n\) and \(y_n\), the construction in Equation~\eqref{eq:synthetic-sufficient-construction} satisfies
\begin{equation}
  \widetilde X^{\top}\widetilde X=G_n,
  \qquad
  \widetilde X^{\top}\widetilde y=b_n.
  \label{eq:synthetic-sufficient-statistics}
\end{equation}
Consequently, the original demonstrations and the synthetic sufficient demonstrations give the same posterior in Equation~\eqref{eq:linear-posterior}. They also give the same ridge estimate for every \(\lambda>0\):
\begin{equation}
  (G_n+\lambda I)^{-1}b_n
  =
  (\widetilde X^{\top}\widetilde X+\lambda I)^{-1}
  \widetilde X^{\top}\widetilde y.
  \label{eq:ridge-synthetic-sufficient-equivalence}
\end{equation}
\end{theorem}
For this regression model, the two records contain the same information. When numerical precision and parsing are held fixed, any difference under the same continuation procedure arises after encoding and not from different stored information. The minimum-row property and spectral truncation bounds appear in Appendix~\ref{app:analysis-results}. The finite-bit construction below connects the stored precision to posterior perturbation, predictive log loss, and the total number of encoded scalars.

\paragraph{Quantized sufficient statistics}

A finite record stores quantized approximations rather than exact real numbers. Let \(\overline G\succeq0\) and \(\overline b\) approximate \(G_n\) and \(b_n\), and define
\begin{equation}
\begin{aligned}
  A&=V_0^{-1}+\sigma^{-2}G_n,
  &h&=V_0^{-1}m_0+\sigma^{-2}b_n,\\
  \overline A&=V_0^{-1}+\sigma^{-2}\overline G,
  &\overline h&=V_0^{-1}m_0+\sigma^{-2}\overline b,
\end{aligned}
\label{eq:quantized-posterior-objects}
\end{equation}
and let \(V=A^{-1}\), \(m=Vh\), \(\overline V=\overline A^{-1}\), and \(\overline m=\overline V\overline h\).

\begin{theorem}[Stability under statistic quantization]
\label{thm:linear-quantization}
Let \(\alpha=\lambda_{\min}(V_0^{-1})>0\). Suppose
\begin{equation}
  \lVert\overline G-G_n\rVert_{\mathrm{op}}\leq\delta_G,
  \qquad
  \lVert\overline b-b_n\rVert_2\leq\delta_b.
  \label{eq:quantization-errors}
\end{equation}
Then, the covariance and mean perturbations satisfy
\begin{equation}
  \lVert\overline V-V\rVert_{\mathrm{op}}
  \leq
  \frac{\delta_G}{\sigma^2\alpha^2},
  \label{eq:covariance-quantization-bound}
\end{equation}
and
\begin{equation}
  \lVert\overline m-m\rVert_2
  \leq
  \frac{\delta_b}{\sigma^2\alpha}
  +\frac{\delta_G\lVert h\rVert_2}
  {\sigma^2\alpha^2}.
  \label{eq:mean-quantization-bound}
\end{equation}
Consequently, for every query with \(\lVert x\rVert_2\leq L_x\), the predictive mean and variance satisfy
\begin{equation}
\begin{aligned}
  \left|x^{\top}(\overline m-m)\right|
  &\leq L_x\lVert\overline m-m\rVert_2,\\
  \left|x^{\top}(\overline V-V)x\right|
  &\leq
  \frac{L_x^2\delta_G}{\sigma^2\alpha^2}.
\end{aligned}
\label{eq:predictive-quantization-bound}
\end{equation}
\end{theorem}

The theorem applies after projecting a symmetrically quantized Gram matrix onto the positive semidefinite cone. Because \(G_n\) is positive semidefinite, this projection does not increase its Frobenius distance from the truth.

\begin{corollary}[Predictive log loss under finite precision]
\label{cor:linear-predictive-kl}
Let \((\mu_x,v_x)\) and \((\overline\mu_x,\overline v_x)\) denote the posterior predictive mean and variance obtained from \((G_n,b_n)\) and \((\overline G,\overline b)\), respectively, at a query with \(\lVert x\rVert_2\leq L_x\). Define
\begin{equation}
\begin{aligned}
  \varepsilon_{\mu}
  &=L_x\left(
    \frac{\delta_b}{\sigma^2\alpha}
    +\frac{\delta_G\lVert h\rVert_2}{\sigma^2\alpha^2}
  \right),\\
  \varepsilon_v
  &=\frac{L_x^2\delta_G}{\sigma^2\alpha^2}.
\end{aligned}
\label{eq:predictive-kl-errors}
\end{equation}
If \(\varepsilon_v\leq\sigma^2/2\), then
\begin{equation}
  \KL\!\left(
    \mathcal N(\mu_x,v_x)
    \mathbin\Vert
    \mathcal N(\overline\mu_x,\overline v_x)
  \right)
  \leq
  \frac{\varepsilon_{\mu}^2}{2\sigma^2}
  +\frac{\varepsilon_v^2}{\sigma^4}.
  \label{eq:predictive-kl-quantization}
\end{equation}
\end{corollary}

\begin{corollary}[A linear handover stored with finitely many bits]
\label{cor:linear-bit-budget}
Suppose \(\lVert x_i\rVert_2\leq L\) and \(|y_i|\leq B_y\). Quantize the unique entries of \(G_n\) with step \(q_G\), quantize \(b_n\) with step \(q_b\), and project the matrix to the positive semidefinite cone. The resulting record uses at most
\begin{equation}
\begin{aligned}
 B_{\mathrm{lin}}
 \leq{}&
 \frac{d(d+1)}{2}
 \left\lceil
 \log_2\left(1+\frac{2nL^2}{q_G}\right)
 \right\rceil\\
 &+d
 \left\lceil
 \log_2\left(1+\frac{2nLB_y}{q_b}\right)
 \right\rceil
 +O(d^2)
 \label{eq:linear-bit-budget}
\end{aligned}
\end{equation}
bits, and Theorem~\ref{thm:linear-quantization} holds with
\begin{equation}
  \delta_G\leq q_G d,
  \qquad
  \delta_b\leq q_b\sqrt d.
  \label{eq:linear-entrywise-errors}
\end{equation}
\end{corollary}

Corollary~\ref{cor:linear-bit-budget} connects exact real-valued sufficiency to a finite code and explains why theoretical bits, serialized bytes, and tokenizer tokens should be reported separately.

\begin{corollary}[Finite-bit predictive rate]
\label{cor:linear-bit-kl-rate}
Under the assumptions of Corollary~\ref{cor:linear-bit-budget}, let
\begin{equation}
  p=\frac{d(d+1)}{2}+d.
  \label{eq:linear-state-dimension}
\end{equation}
For fixed \((n,d,L,B_y,V_0,m_0,\sigma^2,L_x)\), there are constants \(B_0\) and \(K\) such that every \(B\geq B_0\) admits a fixed-length handover of at most \(B\) bits satisfying
\begin{equation}
  \sup_{\lVert x\rVert_2\leq L_x}
  \KL\!\left(
    \mathcal N(\mu_x,v_x)
    \mathbin\Vert
    \mathcal N(\overline\mu_x,\overline v_x)
  \right)
  \leq
  K2^{-2(B-B_0)/p}.
  \label{eq:linear-bit-kl-rate}
\end{equation}
\end{corollary}

\subsection{Nonparametric Regression}
\label{sec:nonparametric}

A fixed-dimensional statistic cannot represent every function in a nonparametric class. We consider
\begin{equation}
  Y_i=f(X_i)+\varepsilon_i,
  \qquad X_i\in[0,1]^d,
  \label{eq:nonparametric-model}
\end{equation}
where the observations are independent and identically distributed. The input density \(p\) satisfies
\begin{equation}
  0<p_{\min}\leq p(x)\leq p_{\max}<\infty,
  \label{eq:density-bounds}
\end{equation}
the noise has conditional mean zero, and \(|Y_i|\leq B_y\) almost surely. For \(0<\beta\leq1\), let \(\mathcal H^\beta(L,B_f)\) contain functions satisfying \(|f(x)|\leq B_f\) and
\begin{equation}
  |f(x)-f(x')|
  \leq L\lVert x-x'\rVert_2^\beta
  \label{eq:holder-condition}
\end{equation}
for every \(x,x'\in[0,1]^d\). We measure error by the integrated squared error over the distribution of later inputs:
\begin{equation}
  \lVert\widehat f-f\rVert_{P_X}^2
  =\int_{[0,1]^d}(\widehat f(x)-f(x))^2p(x)\,dx.
  \label{eq:integrated-risk}
\end{equation}

Partition \([0,1]^d\) into \(M=m^d\) equal cubes \(A_1,\ldots,A_M\). For each cell, the writer stores the count and the sum of the responses,
\begin{equation}
  N_j=\sum_{i=1}^{n}\mathbf 1(X_i\in A_j),
  \qquad
  S_j=\sum_{i:X_i\in A_j}Y_i.
  \label{eq:cell-statistics}
\end{equation}
For \(N_j>0\), let \(\overline Y_j=S_j/N_j\), set \(\overline Y_j=0\) when \(N_j=0\), and let \(Q_q\) be a quantizer with \(|Q_q(z)-z|\leq q\). The resulting handover is
\begin{equation}
  H_M=\left((N_j,Q_q(\overline Y_j)):
  j=1,\ldots,M\right).
  \label{eq:cell-handover}
\end{equation}
The decoder uses the stored mean for a later input in a nonempty cell and zero in an empty cell. The following theorem separates the resulting error into approximation within a cell, sampling variation, empty cells, and quantization.

\begin{theorem}[Risk of a handover based on cell statistics]
\label{thm:nonparametric-upper}
Under Equations~\eqref{eq:nonparametric-model}--\eqref{eq:holder-condition}, suppose \(M\leq n\). There are constants \(C_1,C_2,C_3,C_4\), depending only on \((d,\beta,L,B_f,B_y,p_{\min},p_{\max})\), such that
\begin{equation}
\begin{aligned}
  \sup_{f\in\mathcal H^\beta(L,B_f)}
  \Ex_f[\lVert\widehat f_{H_M}-f\rVert_{P_X}^2]
  \leq{}& C_1M^{-2\beta/d}
  +C_2\frac{M}{n}\\
  &+C_3\exp\left(-\frac{np_{\min}}{M}\right)
  +C_4q^2.
  \label{eq:nonparametric-upper}
\end{aligned}
\end{equation}
\end{theorem}
The terms arise from approximating the function by a constant within each cell, sampling variation, empty cells, and quantization. Choosing
\begin{equation}
  M\asymp n^{d/(2\beta+d)},
  \qquad
  q\asymp n^{-\beta/(2\beta+d)}
  \label{eq:nonparametric-optimal-choice}
\end{equation}
gives
\begin{equation}
  \sup_{f\in\mathcal H^\beta(L,B_f)}
  \Ex_f[\lVert\widehat f_{H_M}-f\rVert_{P_X}^2]
  =O\left(n^{-2\beta/(2\beta+d)}\right),
  \label{eq:nonparametric-minimax-rate}
\end{equation}
which matches the standard full-data order for this class. The same rate can be attained with a finite code for the stored counts and means.

\begin{corollary}[Bit budget needed to attain the full-data rate]
\label{cor:nonparametric-bit-budget}
If each count and each quantized nonempty-cell mean is stored with the fixed-length code described in Appendix~\ref{app:proof-nonparametric-bit-budget}, then the handover uses
\begin{equation}
  B_n=O\left(n^{d/(2\beta+d)}\log n\right)
  \label{eq:nonparametric-bit-budget}
\end{equation}
bits and attains the rate in Equation~\eqref{eq:nonparametric-minimax-rate}.
\end{corollary}

For a fixed design distribution, define the minimax risk under a \(B\)-bit handover by
\begin{equation}
  R_{n,B}
  =\inf_{\substack{P_{H\mid\mathcal D_n},D:\
  H\in\mathcal Z_B,\ |\mathcal Z_B|\leq2^B}}
  \sup_{f\in\mathcal H^\beta(L,B_f)}
  \Ex_f[\lVert\widehat f_{H,D}-f\rVert_{P_X}^2],
  \label{eq:minimax-budget-risk}
\end{equation}
where \(\mathcal D_n\) denotes the \(n\) demonstrations, the decoder observes the message and future query, and the encoder and decoder may be randomized.

\begin{corollary}[Upper bound under a memory limit]
\label{cor:nonparametric-budget-upper}
There is a constant \(c_{\mathrm{code}}>0\), depending only on the fixed coding convention and the class parameters, such that the encoder can use any partition with
\begin{equation}
  M\leq M_B
  =\min\left\{
    n,
    \left\lfloor
      \frac{c_{\mathrm{code}}B}{\log_2(n+1)}
    \right\rfloor
  \right\}
  \label{eq:nonparametric-feasible-cells}
\end{equation}
and store the counts and means with \(q^2\lesssim M^{-2\beta/d}\). For budgets with \(M_B\geq1\), the risk satisfies
\begin{equation}
  R_{n,B}
  \lesssim
  \inf_{1\leq M\leq M_B}
  \left(
    M^{-2\beta/d}
    +\frac{M}{n}
    +\exp\left(-\frac{np_{\min}}{M}\right)
  \right),
  \label{eq:nonparametric-budget-upper}
\end{equation}
where the infimum ranges over partitions of the form \(M=m^d\).
\end{corollary}

The encoder need not spend the entire budget on a finer partition. When \(M_B\) is smaller than the unconstrained choice in Equation~\eqref{eq:nonparametric-optimal-choice}, the largest feasible partition gives the predicted rate, up to coding logarithms, in the range where the memory budget is the binding constraint. Once the budget can encode that unconstrained choice, the encoder keeps the same partition and remains at the sample floor instead of increasing variance by using every available bit.

To obtain a matching lower bound, specialize Equation~\eqref{eq:minimax-budget-risk} to uniform \(X_i\) on \([0,1]^d\) and a bounded binary-response submodel. Conditional on \(X_i=x\), let \(Y_i\in\{-B_y,B_y\}\) satisfy
\begin{equation}
  \Pr(Y_i=B_y\mid X_i=x)
  =\frac{1}{2}\left(1+\frac{f(x)}{B_y}\right),
  \label{eq:bounded-binary-submodel}
\end{equation}
where the packing amplitude is chosen so that \(|f(x)|\leq B_y/2\). This choice gives \(\Ex[Y_i\mid X_i=x]=f(x)\), and the submodel obeys the bounded-outcome assumptions used in the upper bound.

\begin{theorem}[Lower bounds from the sample size and memory limit]
\label{thm:nonparametric-lower}
There is a constant \(c>0\), depending only on \((d,\beta,L,B_f,B_y)\), such that
\begin{equation}
  R_{n,B}
  \geq
  c\max\left\{
    n^{-2\beta/(2\beta+d)},
    (B+1)^{-2\beta/d}
  \right\}.
  \label{eq:nonparametric-lower}
\end{equation}
\end{theorem}
The first term is the error that remains even when the decoder receives all observations, and the second is caused by the finite number of possible handovers. Combined with the upper bound, the theorem shows that the required memory depends on the chosen error level and the function class, not only on the number of observations. Appendix~\ref{app:analysis-results} states extensions to smoother functions and to functions known to depend on fewer coordinates.

\section{Implications for Evaluating Handover Records}
\label{sec:evaluation-protocol}

Comparing arbitrary summaries does not identify the theoretical quantities defined above. A controlled study must specify what the writer knows at the boundary, the distribution of later inputs, every channel through which earlier information remains available, the serialization shown to the continuation procedure, and the task loss.

\subsection{Ideal-Decoder Calibration}

A comparison of writers holds the continuation rule, later inputs, and loss fixed. Its observed effect is generally end-to-end because it reflects both the information carried by each record and the rule's ability to interpret the resulting representation. The three terms in Equation~\eqref{eq:three-loss-decomposition} can be separated only when the ideal risk of each record is known. In the Gaussian setting of Section~\ref{sec:parametric}, the exact posterior calculation supplies that reference, so information loss can be separated from numerical or algorithmic error introduced after the boundary.

The full earlier context and a record with no earlier information are reference conditions, not universal upper and lower bounds for an arbitrary fixed model. A task may contain predictive information in \(T\) or \(X\). A long input may also make a particular continuation procedure less reliable. The theoretical comparison therefore uses excess risk relative to the ideal decoder under the information available in each condition, not an assumed ordering of prompt formats.

\subsection{Information-Equivalent Records}

For a handover record \(H\), define
\begin{equation}
  R^{\star}(H)
  =\inf_D\Ex[\ell(D(H,X,T),Y)],
  \label{eq:record-ideal-risk}
\end{equation}
where the infimum is over continuation rules with the output space and loss fixed in the comparison. Presentation comparisons are controlled only when the information relation between their records is known.

\begin{proposition}[Ideal-risk ordering under deterministic representations]
\label{prop:serialization-ordering}
Let \(H_1\) and \(H_2\) be evaluated with the same \((X,Y,T)\), output space, and loss. If \(H_2=s(H_1,T)\) almost surely for a deterministic map \(s\), then
\begin{equation}
  R^{\star}(H_1)\leq R^{\star}(H_2).
  \label{eq:serialization-risk-ordering}
\end{equation}
If there is also a deterministic map \(r\) such that \(H_1=r(H_2,T)\) almost surely, then the two ideal risks are equal.
\end{proposition}

When the deterministic maps exist in both directions, a fixed continuation rule can still incur different losses because its gap from the common ideal risk may differ across the two representations.

\begin{corollary}[Representation-gap identity]
\label{cor:representation-gap}
Suppose the two deterministic maps in Proposition~\ref{prop:serialization-ordering} exist. For a fixed continuation rule \(D\), write
\begin{equation}
  R(D;H_j)=\Ex[\ell(D(H_j,X,T),Y)]
  \label{eq:fixed-decoder-record-risk}
\end{equation}
for \(j\in\{1,2\}\). Then, we have
\begin{equation}
\begin{aligned}
  R(D;H_1)-R(D;H_2)
  ={}&\left(R(D;H_1)-R^{\star}(H_1)\right)\\
  &-\left(R(D;H_2)-R^{\star}(H_2)\right).
  \label{eq:representation-gap-identity}
\end{aligned}
\end{equation}
\end{corollary}
The identity follows by subtracting the equal ideal risks. It rules out unequal task information as the source of the observed difference but does not identify why the fixed continuation procedure has a different gap from the ideal decoder across the two representations.

In the Gaussian setting, the sufficient-statistic record and the exact synthetic sufficient demonstrations both determine \((G_n,b_n)\). Theorem~\ref{thm:synthetic-sufficient-equivalence} and Proposition~\ref{prop:serialization-ordering} therefore give them the same ideal risk. A truncated or otherwise lossy representation must instead be compared with the exact state, because its risk combines information loss with any additional decoder gap.

\subsection{Memory Accounting}

A short prompt does not imply a small handover when earlier information remains in files, indexes, or databases. A controlled comparison should report the active record, external storage, and content returned to the continuation procedure separately, then use their total when comparing self-contained methods. Serialized bytes and tokenizer tokens are also distinct from the information budget in Equation~\eqref{eq:information-distortion}.

A memory comparison should state the task distribution, the unit in which the budget is measured, and every retained channel. When several budgets are considered, the object of interest is the resulting risk curve rather than one arbitrary operating point.

\section{Discussion}
\label{sec:discussion}

The formulation, handover construction, and statistical results separate three properties that are often combined in a single summary. The record may omit task-relevant information, the memory limit may make some loss unavoidable, and the continuation procedure may fail to attain the risk permitted by the information it receives.

\subsection{What Makes a Handover Better}

The proposed method records what the unfinished task still requires, with decisions and constraints written directly because changing or omitting one can change the next permitted action. Repeated observations are shortened only when the task supplies a theorem or an approximation bound that justifies the replacement. An exceptional observation remains in the record when it determines the next step. The task and its scoring rule therefore determine what the writer may shorten.

Memory size cannot be interpreted independently of the continuation task. The Gaussian example has a finite-dimensional sufficient state whose dimension is independent of the number of demonstrations. The finite-precision results connect code length to predictive log loss. In the nonparametric example, the record must grow with the target accuracy until the sample-size floor is reached. Proposition~\ref{prop:unknown-query-cost} identifies another source of cost: a writer that commits before the realized query may need to retain information about several possible downstream uses that a query-aware encoder could discard.

The representation also matters after the state has been selected. Proposition~\ref{prop:serialization-ordering} identifies when two records have the same ideal risk, and Corollary~\ref{cor:representation-gap} assigns any remaining difference under one fixed continuation procedure to its gap from the ideal decoder. This separation prevents a failure to parse or use a representation from being described as information loss by the writer.

\subsection{Relation to Current Memory Systems}

Compaction shortens the prompt, persistent memory inserts selected records into later sessions, and retrieval returns content from an external source. A KV cache serves a different role by reducing repeated computation without recording the task. These mechanisms can help an application continue, but they do not determine what the next decision requires.

External records change the meaning of a memory limit because a short prompt does not limit the amount stored in files or databases. The formulation therefore counts prompt content, external storage, and returned content as parts of the handover whenever they derive from the earlier session.

\subsection{Scope of the Theory}

The general results apply to the joint distribution of earlier context, later input, target, and task information. The regression models furnish solvable instances in which the relevant state and the ideal decoder can be characterized, yielding exact and approximate memory requirements for those tasks.

Predictive sufficiency is relative to the later-input law. A change in that law can alter which information is sufficient, and an adaptive trajectory requires a trajectory-level loss because earlier actions change later observations. The results here concern the session boundary under the continuation law.

\section{Conclusion}
\label{sec:conclusion}

A session handover succeeds when the information passed to the new session supports the continuation required by the task. By modeling the writer, the prompt built from the record, the continuation procedure, and the task score separately, we distinguish reproduction of earlier text from preservation of task information and separate information omitted by the writer from loss introduced after the record has been supplied.

The proposed method records decisions and constraints directly, shortens repeated evidence only when the relation between the replacement and task loss is explicit, and keeps observations that cannot be replaced. Predictive equivalence characterizes the coarsest deterministic sufficient state under the exogeneity condition of Proposition~\ref{prop:coarsest-handover}, and the memory analysis shows how the budget and the timing of the writer constrain attainable risk. Gaussian regression gives an exact finite-dimensional handover and finite-bit predictive bounds, and nonparametric regression gives an achievable memory--risk relation and a distinct memory floor. These results establish a theoretical basis for constructing and comparing handover records across session boundaries.

\bibliography{arXiv2.bbl}

\bibliographystyle{tmlr}

\clearpage

\appendix

\section{Existing Memory Mechanisms and Related Work}
\label{app:memory-systems}

This appendix places session handover among prompt compression, persistent memory, external records, runtime caches, and in-context learning theory. The distinctions concern both when information is written and which task-relative property it is meant to preserve.

\subsection{Boundary Events and Storage Channels}

Context carryover asks whether information from an earlier turn remains available later. A handover occurs when a new session or another agent continues an unfinished task, and the handover record is the information selected for that boundary. Context management determines what enters one model call, while memory management also includes writing, storing, updating, and retrieving information across calls.

Earlier information can remain in the active prompt, a compacted replacement, persistent instructions, an external file or database, or a retrieval index. Retrieval-augmented generation places selected external records back into the model input \citep{Lewis2020retrievalaugmented}. These mechanisms can carry a handover, but none determines by itself whether the retained content is sufficient for the continuation task.

A KV cache belongs to model execution, not to the task record, and stores attention keys and values so that earlier token states need not be recomputed. PagedAttention and distributed KV-cache systems reduce serving cost and data movement \citep{Kwon2023efficientmemory,Qin2025mooncaketrading}, but cached activations do not form an auditable account of the task state. We therefore separate computational cache cost from the information retained in \(H\).

\subsection{Prompt Compression and Source Coding}

Prompt compression shortens an input to preserve a model output. Rate--distortion formulations explain how a representation limit constrains the error after compression and why an encoder that observes the realized query can have an advantage \citep{Nagle2024fundamentallimit,Colaco2026keepforget}. A handover writer has a different information pattern: it knows the task and earlier context when it writes, but the continuation procedure observes the realized later input afterward.

The information bottleneck seeks a compressed representation that retains information about a relevant variable \citep{Tishby2000theinformation}. Indirect source coding studies an encoder that observes a proxy for the target, Wyner--Ziv coding studies side information at the decoder, and Blackwell's comparison of experiments orders information sources by the decisions they support \citep{Wolf1970transmissionof,Wyner1976therate,Blackwell1953equivalentcomparisons}. The formulation in Section~\ref{sec:formulation} uses these classical tools to define the state needed for one continuation task and to place the writer before the realized later query.

Repeated compaction and structured eviction address related operational problems. Parallel Context Compaction studies repeated summarization, while Structured Context Eviction uses typed records and dependencies to determine what may be removed \citep{Cim2026parallelcontext,Semenov2026compactionstructured}. The present analysis does not prescribe a particular compaction algorithm but supplies a task-relative criterion against which a resulting record can be judged.

\subsection{In-Context Learning and Sufficient State}

Theory of in-context learning shows that transformers can implement regression and other learning procedures from examples in the prompt \citep{Bai2023transformersas,VanOswald2023transformerslearn}. Other work derives prediction-error rates for broader function classes and studies how pretrained transformers exploit low-dimensional task structure \citep{Kim2024transformerslearn,Kim2024transformersare,Oko2024pretrainedtransformer}. Our starting point is the induced prediction rule after the demonstrations have been processed, and a handover must retain the task-relevant part of that state across the session boundary.

The Gaussian sufficient statistics and synthetic sufficient demonstrations in Section~\ref{sec:parametric} make the distinction explicit because they determine the same posterior distribution and are therefore equivalent for an ideal decoder even though their token-level forms differ. This use of statistical sufficiency complements work that studies which algorithm a transformer can implement from raw examples.

\section{Additional Details for the Handover Method}
\label{app:handover-details}

This appendix gives additional details for the writer, explaining how selected observations may remain outside the prompt and how comparisons keep the information content of different presentations fixed.

\subsection{Selected Observations Stored outside the Prompt}

Equation~\eqref{eq:selected-external-records} places selected original observations in external records and leaves their identifiers in the prompt. Because both parts count toward the handover, the size limit includes the identifiers, metadata, stored observations, and any content returned to the model.

The proposed method does not retain the entire earlier history outside the prompt. It stores only observations whose effect on the task is not preserved by the shorter record, and the selection rule and resulting storage cost are part of the writer.

\subsection{Serialization and Deterministic Checks}

The same state can be presented as concise prose, structured fields, numerical statistics, or synthetic sufficient demonstrations, and two forms have the same information content when each determines that state. When the forms do not determine each other, we order them only if one can be calculated deterministically from the other.

Proposition~\ref{prop:serialization-ordering} and Corollary~\ref{cor:representation-gap} are stated in Section~\ref{sec:evaluation-protocol}. In the Gaussian setting, direct statistics and exact synthetic sufficient demonstrations determine the same \((G_n,b_n)\), so they have the same ideal risk. More generally, deterministic processing cannot improve the best attainable risk, and two forms that determine each other remain equivalent at the ideal-decoder level.

The writer first records the events that affect later decisions. It then calculates any statistic required by the task, selects the original examples that must remain, and divides the available space among the fields. Before the record is used, deterministic checks verify its schema, numerical dimensions, reference targets, and size, with the log attributing these checks to the writer.

For a long-running task, the application can update exact fields when a constraint is added or a decision is made, which avoids asking the model to reconstruct every commitment only when the context is almost full. The theory also covers a writer that constructs the record only at the boundary.

\section{Additional Results for the General Formulation and Statistical Analysis}
\label{app:analysis-results}

This appendix collects supporting results for Sections~\ref{sec:formulation} and~\ref{sec:statistical-analysis}.

\subsection{Response Distributions under the Two Prompts}

For the formulation with one model call in Section~\ref{sec:formulation}, fix the two prompts in Equation~\eqref{eq:full-handover-prompts}. Let
\begin{equation}
  P=\Pi_{\theta,\delta}(\cdot\mid p_{\mathrm{full}}),
  \qquad
  Q=\Pi_{\theta,\delta}(\cdot\mid p_H)
  \label{eq:response-pq}
\end{equation}
be distributions on padded response sequences of length \(L_{\mathrm{out}}\).

\begin{proposition}[Autoregressive decomposition of response divergence]
\label{prop:autoregressive-response-kl}
Suppose that \(P\) and \(Q\) admit autoregressive factorizations, that \(P\) is absolutely continuous with respect to \(Q\), and that \(P_m(\cdot\mid Z_{1:m-1})\) and \(Q_m(\cdot\mid Z_{1:m-1})\) denote their next-token conditional distributions. The chain rule then gives
\begin{equation}
\begin{aligned}
  \KL(P\mathbin\Vert Q)
  =\sum_{m=1}^{L_{\mathrm{out}}}
  \Ex_{Z_{1:m-1}\sim P}
  \left[
    \KL\left(
      P_m(\cdot\mid Z_{1:m-1})
      \mathbin\Vert
      Q_m(\cdot\mid Z_{1:m-1})
    \right)
  \right].
  \label{eq:response-kl-chain}
\end{aligned}
\end{equation}
\end{proposition}

The proposition applies the standard chain rule for relative entropy to two prompts of the same model. The resulting decomposition shows at which token positions the two generations begin to differ, although agreement with the model given the full context does not establish that either response is correct for the task.

For an interactive agent, let \(O_0=X\), let \(A_k\) be the action at step \(k\), and let \(O_{k+1}\) be the next environment observation. With \(\tau=(O_0,A_0,O_1,\ldots,A_{K-1},O_K)\), an explicit version of Equation~\eqref{eq:trajectory-distribution} is
\begin{equation}
\begin{aligned}
  &P_{E,\theta,\delta,Q}(d\tau\mid C,T,X)\\
  &\quad=\int P_E(dh\mid C,T)
  \prod_{k=0}^{K-1}
  \pi_{\theta,\delta}
  (dA_k\mid h,X,T,\tau_{<k})
  Q(dO_{k+1}\mid X,T,\tau_{\leq k}).
  \label{eq:interactive-trajectory-factorization}
\end{aligned}
\end{equation}
The environment rule \(Q\) may be deterministic, and the factorization states where the handover enters the trajectory without assuming that the actions leave later observations fixed.

\subsection{Sequential Prediction and Changes in the Distribution of Later Inputs}

For a finite target space, a bounded proper score gives another exact measure of the posterior information lost at the boundary.

\begin{proposition}[Posterior distortion under the Brier score]
\label{prop:brier-posterior-distortion}
Let \(\mathcal Y=\{1,\ldots,K\}\), and let a prediction be a probability vector \(q\) on \(\mathcal Y\). Define the Brier loss by
\begin{equation}
  \ell_{\mathrm{Br}}(q,Y)
  =\sum_{y\in\mathcal Y}
  \left(q_y-\mathbf 1\{Y=y\}\right)^2.
  \label{eq:brier-loss}
\end{equation}
Under Proposition~\ref{prop:log-loss-identity}, let
\begin{equation}
  p_C=P(Y=\cdot\mid C,X,T),
  \qquad
  p_H=P(Y=\cdot\mid H,X,T).
  \label{eq:brier-posteriors}
\end{equation}
Then, the Bayes risks satisfy
\begin{equation}
  R_{\mathrm{Br}}^{\star}(H)
  -R_{\mathrm{Br}}^{\star}(C)
  =\Ex\left[\lVert p_C-p_H\rVert_2^2\right].
  \label{eq:brier-posterior-distortion}
\end{equation}
Consequently, the Brier-risk increase is zero if and only if the handover is predictively sufficient for the finite target.
\end{proposition}

A freely acting trajectory requires more than sufficiency for one prediction because each action can change what the model observes next. A fixed replay avoids this problem and retains a sequence of decisions.

\begin{theorem}[Log-loss decomposition over a fixed horizon]
\label{thm:fixed-horizon-log-loss}
Let \(K\geq1\), let \(H\) be generated from \((C,T)\), and suppose a fixed evaluation law generates \((X_{1:K},Y_{1:K})\) without depending on predictions made by the compared decoders. Assume that \(H\) and \((X_{1:K},Y_{1:K})\) are conditionally independent given \((C,T)\). At step \(k\), the handover predictor observes \((H,T,X_{1:k},Y_{<k})\), and the predictor given the full context observes \((C,T,X_{1:k},Y_{<k})\). Let their cumulative Bayes log risks be the sums of the corresponding conditional entropies. Then, the risk difference decomposes as
\begin{equation}
\begin{aligned}
  R_{\log,K}^{\star}(H)
  -R_{\log,K}^{\star}(C)
  =\sum_{k=1}^{K}
  I\left(
    Y_k;C
    \mid H,T,X_{1:k},Y_{<k}
  \right).
  \label{eq:fixed-horizon-log-loss}
\end{aligned}
\end{equation}
\end{theorem}

Theorem~\ref{thm:fixed-horizon-log-loss} applies to logged replay, prediction at fixed checkpoints, and other protocols that give every condition the same sequence after the handover. Trajectories in which different policies choose different actions and receive different later observations fall outside this result.

An application may rewrite its compacted record several times. Theorem~\ref{thm:repeated-handover} separates repeated rewriting from the case in which a later session receives new evidence about the task.

\begin{theorem}[Repeated handover without new evidence]
\label{thm:repeated-handover}
Let \(H_0=C\). For \(j=1,\ldots,m\), suppose \(H_j\) is generated from \(H_{j-1}\) and \(T\), and suppose that, conditional on \(T\),
\begin{equation}
  (C,X,Y)\longrightarrow H_{j-1}\longrightarrow H_j
  \label{eq:repeated-handover-markov}
\end{equation}
holds as a Markov relation. If the Bayes log risks below are finite, then the following identity holds:
\begin{equation}
\begin{aligned}
  R_{\log}^{\star}(H_m)-R_{\log}^{\star}(C)
  =\sum_{j=1}^{m}
  I(Y;H_{j-1}\mid H_j,X,T).
\end{aligned}
\label{eq:repeated-handover-decomposition}
\end{equation}
Consequently, the ideal log risk is nondecreasing across the rewrites.
\end{theorem}

Theorem~\ref{thm:repeated-handover} applies when each stage only rewrites the state it received. A later stage can improve prediction when it obtains new current-state observations or tool results, because those observations add information that was unavailable to the earlier record.

Predictive sufficiency depends on the distribution of later queries. The next result considers a change in that distribution when the new queries reveal no additional information about the context available before the handover.

\begin{proposition}[A shift in the distribution of exogenous queries]
\label{prop:query-shift}
Fix \(T=t\), and suppose \(X\) is independent of \((C,H)\) under both a reference query distribution \(\mu_t\) and an evaluation query distribution \(\nu_t\). Hold fixed the joint law of \((C,H)\) and the conditional target kernel \(P(Y\mid C,X,T=t)\). Let \(\Delta_{\mu_t}(H)\) and \(\Delta_{\nu_t}(H)\) be the Bayes log-risk increases caused by using \(H\) instead of \(C\) under the two query distributions. If \(\nu_t\) is absolutely continuous with respect to \(\mu_t\) and
\begin{equation}
  \operatorname*{ess\,sup}_{x}
  \frac{d\nu_t}{d\mu_t}(x)
  \leq\kappa,
  \label{eq:query-density-ratio}
\end{equation}
then the evaluation-distribution increase satisfies
\begin{equation}
  0\leq\Delta_{\nu_t}(H)
  \leq\kappa\Delta_{\mu_t}(H).
  \label{eq:query-shift-bound}
\end{equation}
\end{proposition}

The coverage condition is essential: if \(\nu_t\) assigns mass to queries outside the support of \(\mu_t\), a record can be sufficient under \(\mu_t\) and lose arbitrary target information on the new queries. The bound therefore applies only when the later-input distribution used for evaluation satisfies the coverage condition in Equation~\eqref{eq:query-density-ratio}.

Equation~\eqref{eq:bounded-loss-information} applies to a particular task, so a record can be sufficient for choosing the next admissible action yet remain insufficient for reconstructing a detailed narrative. Its proof appears in Appendix~\ref{app:general-proofs}.

\subsection{Additional Information-Budget Results}

When external records are available, we write all information passed through the handover as the pair
\begin{equation}
  H=(V,M),
  \qquad
  \len(V)\leq B_{\mathrm{act}},
  \label{eq:active-external-record-channel}
\end{equation}
where \(V\) is the record placed in the prompt and \(M\) contains information retained outside it. The limit \(B_{\mathrm{act}}\) controls \(V\). The information constraint in Equation~\eqref{eq:information-distortion} applies to the full pair \((V,M)\). Any application of the formulation must account separately for bytes stored outside the prompt, access requests, and content returned to the continuation procedure.

\begin{proposition}[Structure of the information-budget curve]
\label{prop:information-budget-curve}
Assume that the information quantities below are finite and measured in the same units. Then, \(D_{\mathrm{info}}(B)\) is nonincreasing and convex in \(B\), and it satisfies
\begin{equation}
  0\leq D_{\mathrm{info}}(B)
  \leq I(Y;C\mid X,T).
  \label{eq:information-budget-basic-bounds}
\end{equation}
At zero budget, \(D_{\mathrm{info}}(0)=I(Y;C\mid X,T)\). If \(C\) is discrete and \(B\geq H(C\mid T)\), then \(D_{\mathrm{info}}(B)=0\). If the future query is exogenous, so \(X\) and \(C\) are conditionally independent given \(T\), then it also holds that
\begin{equation}
  D_{\mathrm{info}}(B)
  \geq
  \max\left\{
    0,
    I(Y;C\mid X,T)-B
  \right\}.
  \label{eq:information-budget-linear-lower}
\end{equation}
\end{proposition}

Proposition~\ref{prop:information-budget-curve} gives a one-episode trade-off between stored information and discarded target information. Operational rates for repeated episodes depend on the chosen block-coding model and code construction.

In the main setting, the writer knows \(T\) but does not observe the realized \(X\). A generic compressor may ignore task information at the boundary, whereas a query-aware compressor sees the specific downstream query before choosing what to retain. Handover lies between these cases because the writer knows the task and local goal but commits before the exact later query is known. An encoder that sees \(X\) can discard information unused by the realized query, but the handover writer must account for the distribution of later queries fixed by the task.

Two further bounds clarify what a finite channel can and cannot preserve. The first gives an explicit code when the sufficient state has continuous and discrete parts.

\begin{proposition}[Description length of a mixed sufficient state]
\label{prop:mixed-state-description}
Suppose a predictively sufficient state is \(S=(\theta,m)\), where \(\theta\in[0,1]^d\) and \(m\in\prod_{j=1}^{J}\mathcal A_j\) with finite \(\mathcal A_j\). Assume \(L_S>0\) and that replacing \(\theta\) by \(\theta'\) with \(m\) fixed increases the ideal decoder's risk by at most \(L_S\lVert\theta-\theta'\rVert_2\). For every \(\varepsilon>0\), there is a fixed-length handover whose excess risk relative to the exact state is at most \(\varepsilon\) and whose length satisfies
\begin{equation}
  B_{\mathrm{mix}}(\varepsilon)
  \leq
  d\left\lceil
    \log_2\left(1+\frac{L_S\sqrt d}{\varepsilon}\right)
  \right\rceil
  +\sum_{j=1}^{J}
  \left\lceil\log_2|\mathcal A_j|\right\rceil.
  \label{eq:mixed-state-description}
\end{equation}
\end{proposition}

Under this assumption, the discrete part must be represented exactly and may encode a finite decision status, a task mode, or another state component whose alternatives cannot be averaged without changing the decision problem. The continuous term depends on target accuracy rather than on transcript length.

The next result gives a regret lower bound when distinct histories require separated continuation policies.

\begin{theorem}[Lower bound from separated predictive states]
\label{thm:state-packing}
Let \(J\) be uniform on \(\{1,\ldots,N\}\), with \(N\geq2\), and let \(J\) index histories available before the handover under a fixed task value \(t\). The encoder writes \(H\) before the future observation \(X\) is drawn, with \(X\) independent of \((J,H)\) given \(T=t\). A continuation policy \(\pi\) maps a realized \(X\) to an action distribution, with \(L_j(\pi)\) denoting its expected loss in instance \(j\) and \(L_j^\star=\inf_\pi L_j(\pi)\) denoting the optimum. Suppose that, for some \(\delta>0\), every policy satisfies
\begin{equation}
  \#\left\{j:
    L_j(\pi)<L_j^\star+\delta
  \right\}\leq1.
  \label{eq:policy-packing-condition}
\end{equation}
For every decoder \(D\), let the policy induced by message \(h\) be the conditional action distribution \(\pi_h(\cdot\mid x)=P_D(A\in\cdot\mid h,x,t)\), which integrates over any private randomization. If the handover satisfies \(I(J;H\mid T=t)\leq B\) nats, then every writer and decoder satisfies
\begin{equation}
  \Ex[\Regret]
  \geq
  \delta\max\left\{
    0,
    1-\frac{B+\log 2}{\log N}
  \right\}.
  \label{eq:state-packing-lower}
\end{equation}
\end{theorem}

The packing condition states the decision separation needed for an information lower bound and can be enforced in a synthetic continuation family by assigning mutually incompatible near-optimal policies to the histories. Additional observations may reveal the history and violate the condition. The theorem therefore applies to task families that satisfy the separation condition.

\subsection{Additional Results for Parametric Regression}

\begin{corollary}[Minimum number of exact synthetic sufficient demonstrations]
\label{cor:synthetic-sufficient-rank-minimal}
Let \(Z\in\mathbb R^{s\times d}\) be the input matrix of any synthetic sufficient demonstration set that satisfies \(Z^{\top}Z=G_n\). Then, the row count satisfies
\begin{equation}
  s\geq\rank(G_n).
  \label{eq:synthetic-sufficient-rank-minimal}
\end{equation}
Consequently, the construction in Equation~\eqref{eq:synthetic-sufficient-construction} uses the minimum possible number of real-valued synthetic sufficient demonstrations among exact Gram-matrix representations.
\end{corollary}

The theorem and corollary fix the information content and the number of examples in two presentation formats: a matrix and an exact synthetic sufficient demonstration set contain the same sufficient state. Subject to numerical precision and prompt parsing, any difference when the same fixed model receives the two forms therefore arises downstream of the stored information.

\paragraph{Small eigenvalues and truncated synthetic sufficient demonstrations}

Because the exact construction can produce large synthetic outputs when \(G_n\) has very small positive eigenvalues, consider a truncated construction for which \(U_\tau\) contains the eigenvectors whose eigenvalues exceed \(\tau>0\) and the retained statistics are
\begin{equation}
  G_{n,\tau}=U_\tau\Lambda_\tau U_\tau^{\top},
  \qquad
  b_{n,\tau}=U_\tau U_\tau^{\top}b_n.
  \label{eq:truncated-statistics}
\end{equation}
The corresponding synthetic sufficient demonstrations use Equation~\eqref{eq:synthetic-sufficient-construction} with \((U_\tau,\Lambda_\tau,b_{n,\tau})\).

\begin{proposition}[Ridge error after spectral truncation]
\label{prop:truncated-ridge}
Let
\begin{equation}
  \widehat\beta_\lambda=(G_n+\lambda I)^{-1}b_n,
  \qquad
  \widehat\beta_{\lambda,\tau}
  =(G_{n,\tau}+\lambda I)^{-1}b_{n,\tau}.
  \label{eq:truncated-ridge-estimators}
\end{equation}
Then, for every \(\lambda>0\), the truncation error satisfies
\begin{equation}
  \lVert\widehat\beta_\lambda-
  \widehat\beta_{\lambda,\tau}\rVert_2
  \leq
  \frac{\sqrt{\tau}}{\lambda}\lVert y_n\rVert_2.
  \label{eq:truncated-ridge-bound}
\end{equation}
\end{proposition}

The bound shows the trade-off between exact preservation and the range of the stored numbers. A truncated record must therefore state the threshold and the information discarded by the truncation, and it cannot be described as exact.

\subsection{Higher Smoothness and Intrinsic Dimension}

For \(\beta>1\), a piecewise constant handover does not attain the full H\"older rate, so the encoder must store local polynomial moments through order \(p=\lfloor\beta\rfloor\). Each cell contains
\begin{equation}
  K_{d,p}=\binom{d+p}{d}
  \label{eq:local-polynomial-count}
\end{equation}
coefficients. The primary theorem covers \(0<\beta\leq1\), but the local-polynomial extension requires a separate proof.

The ambient dimension can also overstate memory when the task distribution lies on a known lower-dimensional subspace. In that case, the cell construction can use coordinates on that subspace and replace \(d\) by its dimension. Existing ICL theory shows that pretrained transformers can exploit nonparametric structure and low-dimensional task families under suitable training distributions \citep{Kim2024transformersare,Oko2024pretrainedtransformer}. The rate derived here applies to the regression setting above when the relevant coordinates are known to the encoder.

\section{Proofs for General Handover Results}
\label{app:general-proofs}

This appendix proves the general results in the order in which they are used. Each subsection proves the result named in its heading and uses only the assumptions in that result.

\subsection{Proof of Proposition~\ref{prop:autoregressive-response-kl}}

The autoregressive factorizations give
\begin{equation}
  \log\frac{P(Z_{1:L_{\mathrm{out}}})}
  {Q(Z_{1:L_{\mathrm{out}}})}
  =\sum_{m=1}^{L_{\mathrm{out}}}
  \log\frac{
    P_m(Z_m\mid Z_{1:m-1})
  }{
    Q_m(Z_m\mid Z_{1:m-1})
  }.
  \label{eq:response-kl-log-ratio}
\end{equation}
Taking expectation under \(P\) and applying the tower property to each summand gives
\begin{equation}
\begin{aligned}
  \KL(P\mathbin\Vert Q)
  &=\sum_{m=1}^{L_{\mathrm{out}}}
  \Ex_{Z_{1:m-1}\sim P}
  \left[
    \Ex_{Z_m\sim P_m(\cdot\mid Z_{1:m-1})}
    \left[
      \log\frac{
        P_m(Z_m\mid Z_{1:m-1})
      }{
        Q_m(Z_m\mid Z_{1:m-1})
      }
    \right]
  \right]\\
  &=\sum_{m=1}^{L_{\mathrm{out}}}
  \Ex_{Z_{1:m-1}\sim P}
  \left[
    \KL\left(
      P_m(\cdot\mid Z_{1:m-1})
      \mathbin\Vert
      Q_m(\cdot\mid Z_{1:m-1})
    \right)
  \right],
\end{aligned}
\end{equation}
which proves Equation~\eqref{eq:response-kl-chain}.

\subsection{Proof of Proposition~\ref{prop:coarsest-handover}}

Suppose first that \(H\) is predictively sufficient. If two supported contexts \(c\) and \(c'\) satisfy \(E(c,t)=E(c',t)=h\), exogeneity gives both contexts the same future-query law \(\mu_t\). Predictive sufficiency therefore gives
\begin{equation}
  P(Y\mid C=c,X=x,T=t)
  =P(Y\mid H=h,X=x,T=t)
  \label{eq:coarsest-proof-c}
\end{equation}
for \(\mu_t\)-almost every \(x\), and the same equality holds with \(c'\) in place of \(c\). Intersecting the two full-measure sets shows that \(c\sim_t c'\).

Conversely, suppose every encoder fiber lies in one equivalence class. Because \(X\) is independent of \(C\) given \(T=t\), conditioning on \(X\) does not change the mixture weights over contexts within an encoder fiber. Conditional on \((H=h,X=x,T=t)\), the target distribution is a mixture over the supported contexts in that fiber. For \(\mu_t\)-almost every \(x\), every distribution in the mixture is the same by assumption, so the mixture equals that common distribution. Therefore, Equation~\eqref{eq:predictive-sufficiency} holds.

The quotient map groups exactly the contexts in one equivalence class, so the preceding argument proves its sufficiency. If \(E\) is any deterministic sufficient encoder, define \(\rho_t(h)\) as the equivalence class containing the fiber associated with \(h\). The first part shows that this class is well defined, with \(q_t(c)=\rho_t(E(c,t))\) on the support. Thus, the quotient record can be recovered from every deterministic sufficient handover.

\subsection{Proof of Corollary~\ref{cor:predictive-state-count}}

By Proposition~\ref{prop:coarsest-handover}, a deterministic sufficient handover cannot assign the same codeword to two distinct values of \(q_t(C)\), and a fixed-length \(b\)-bit record has at most \(2^b\) codewords, so \(2^b\geq N_t\) and hence \(b\geq\lceil\log_2 N_t\rceil\). Conversely, indexing the \(N_t\) equivalence classes uses \(\lceil\log_2 N_t\rceil\) bits and is predictively sufficient.

\subsection{Proof of Proposition~\ref{prop:recoverability-sufficiency}}

The encoder in Equation~\eqref{eq:encoder-channel} is conditionally independent of \((X,Y)\) given \((C,T)\). This conditional independence gives
\begin{equation}
  P(Y\mid C,H,X,T)
  =P(Y\mid C,X,T).
  \label{eq:recoverability-encoder-markov}
\end{equation}
If \(C=r(H,T)\) holds almost surely, then conditioning on \((H,X,T)\) also determines \(C\). Consequently, the conditional distributions satisfy
\begin{equation}
  P(Y\mid H,X,T)
  =P(Y\mid C,H,X,T),
  \label{eq:recoverability-determines-context}
\end{equation}
which proves predictive sufficiency.

Suppose next that predictive sufficiency holds. The minimum conditional risk at any realized information set is the Bayes envelope of the conditional distribution of \(Y\). Equation~\eqref{eq:predictive-sufficiency} makes the pointwise Bayes envelopes with \((C,X,T)\) and \((H,X,T)\) equal almost surely, so taking expectations proves \(R_{\ell}^{\star}(H)=R_{\ell}^{\star}(C)\).

For a counterexample to the first converse, let \(U\) and \(V\) be independent \(\operatorname{Bernoulli}(1/2)\) variables, and set \(C=(U,V)\), \(H=U\), and \(Y=U\), with \(X\) and \(T\) constant. The target is determined by \(H\), so predictive sufficiency holds, but \(V\) cannot be recovered from \(H\). Exact recovery of one part is also insufficient: if \(O=V\) and \(H=V\), then \(O\) is recoverable but \(Y=U\) is not predictable from the record.

For a decision problem whose Bayes risk is preserved without predictive sufficiency, let \(C\) be equiprobable on \(\{0,1\}\), let \(H\) be constant, and let \(Y\in\{0,1\}\) satisfy \(\Pr(Y=1\mid C=0)=0.8\) and \(\Pr(Y=1\mid C=1)=0.9\). Under zero--one loss, the Bayes action is \(1\) for both values of \(C\), and the average Bayes risk is \(0.15\) with either \(C\) or \(H\). The conditional target distributions nevertheless differ, so predictive sufficiency fails.

\subsection{Proof of Proposition~\ref{prop:state-recovery-sufficiency}}

Predictive sufficiency of \(S=\phi(C,T)\) gives
\begin{equation}
  P(Y\mid C,X,T)=P(Y\mid S,X,T).
  \label{eq:state-sufficiency-proof-one}
\end{equation}
Exact recoverability gives \(S=r(H,T)\) almost surely, so conditioning on \((H,X,T)\) determines \(S\). Because \(H\) is generated from \((C,T)\), conditioning on \(H\) only changes the mixture over contexts that share this recovered value of \(S\). Equation~\eqref{eq:state-sufficiency-proof-one} gives the same target distribution for every context in that mixture. The resulting mixture therefore satisfies
\begin{equation}
  P(Y\mid H,X,T)=P(Y\mid S,X,T).
  \label{eq:state-sufficiency-proof-two}
\end{equation}
Combining the two equalities proves Equation~\eqref{eq:predictive-sufficiency}.

\subsection{Proof of Proposition~\ref{prop:log-loss-identity}}

The Bayes log risk with information \(Z\) is the conditional entropy of \(Y\) given \(Z\). Therefore, we have
\begin{equation}
  R_{\log}^{\star}(H)=H(Y\mid H,X,T)
  \label{eq:log-risk-h}
\end{equation}
and
\begin{equation}
  R_{\log}^{\star}(C)=H(Y\mid C,X,T).
  \label{eq:log-risk-c}
\end{equation}
Because \(H\) is generated from \((C,T)\), the conditional entropy satisfies
\begin{equation}
  H(Y\mid C,H,X,T)=H(Y\mid C,X,T).
  \label{eq:encoder-markov-entropy}
\end{equation}
The definition of conditional mutual information now gives
\begin{equation}
\begin{aligned}
  I(Y;C\mid H,X,T)
  & =H(Y\mid H,X,T)
  -H(Y\mid C,H,X,T)\\
  & =R_{\log}^{\star}(H)-R_{\log}^{\star}(C),
\end{aligned}
\end{equation}
which proves Equation~\eqref{eq:log-loss-identity}.

\subsection{Proof of Proposition~\ref{prop:brier-posterior-distortion}}

For a probability vector \(p\) on \(\mathcal Y\), the conditional Brier risk of a prediction \(q\) is
\begin{equation}
\begin{aligned}
  \Ex[\ell_{\mathrm{Br}}(q,Y)\mid p]
  &=\sum_{y\in\mathcal Y}q_y^2
    -2\sum_{y\in\mathcal Y}q_yp_y+1.
\end{aligned}
\label{eq:brier-conditional-risk}
\end{equation}
The expression is minimized at \(q=p\), where its value is \(1-\lVert p\rVert_2^2\). At this minimizer,
\begin{equation}
\begin{aligned}
  R_{\mathrm{Br}}^{\star}(H)
  -R_{\mathrm{Br}}^{\star}(C)
  &=\Ex[\lVert p_C\rVert_2^2]
    -\Ex[\lVert p_H\rVert_2^2].
\end{aligned}
\label{eq:brier-risk-norms}
\end{equation}
The encoder Markov condition gives \(p_H=\Ex[p_C\mid H,X,T]\). The conditional Pythagorean identity then yields
\begin{equation}
  \Ex[\lVert p_C\rVert_2^2]
  =\Ex[\lVert p_H\rVert_2^2]
  +\Ex[\lVert p_C-p_H\rVert_2^2].
  \label{eq:brier-pythagorean}
\end{equation}
Combining Equations~\eqref{eq:brier-risk-norms} and~\eqref{eq:brier-pythagorean} proves Equation~\eqref{eq:brier-posterior-distortion}. The right-hand side is zero exactly when \(p_C=p_H\) almost surely, which is predictive sufficiency for a finite target space.

\subsection{Proof of Theorem~\ref{thm:fixed-horizon-log-loss}}

Let \(Z_k=(T,X_{1:k},Y_{<k})\). The cumulative Bayes log risks are
\begin{equation}
  R_{\log,K}^{\star}(H)
  =\sum_{k=1}^{K}H(Y_k\mid H,Z_k),
  \qquad
  R_{\log,K}^{\star}(C)
  =\sum_{k=1}^{K}H(Y_k\mid C,Z_k).
  \label{eq:fixed-horizon-risk-entropies}
\end{equation}
The conditional independence assumption implies
\begin{equation}
  H(Y_k\mid C,H,Z_k)=H(Y_k\mid C,Z_k).
  \label{eq:fixed-horizon-markov}
\end{equation}
Consequently, each summand satisfies
\begin{equation}
\begin{aligned}
  H(Y_k\mid H,Z_k)-H(Y_k\mid C,Z_k)
  &=H(Y_k\mid H,Z_k)-H(Y_k\mid C,H,Z_k)\\
  &=I(Y_k;C\mid H,Z_k).
\end{aligned}
\label{eq:fixed-horizon-step}
\end{equation}
Summing Equation~\eqref{eq:fixed-horizon-step} over \(k\) proves Equation~\eqref{eq:fixed-horizon-log-loss}.

\subsection{Proof of Theorem~\ref{thm:repeated-handover}}

For every \(j\), the Markov relation in Equation~\eqref{eq:repeated-handover-markov} implies
\begin{equation}
  H(Y\mid H_{j-1},H_j,X,T)
  =H(Y\mid H_{j-1},X,T).
  \label{eq:repeated-handover-entropy}
\end{equation}
Subtracting the two equalities gives
\begin{equation}
\begin{aligned}
  R_{\log}^{\star}(H_j)-R_{\log}^{\star}(H_{j-1})
  & =H(Y\mid H_j,X,T)-H(Y\mid H_{j-1},X,T)\\
  & =I(Y;H_{j-1}\mid H_j,X,T).
\end{aligned}
\label{eq:repeated-handover-increment}
\end{equation}
Summing Equation~\eqref{eq:repeated-handover-increment} from \(j=1\) to \(m\) telescopes to Equation~\eqref{eq:repeated-handover-decomposition}. Every conditional mutual information in the sum is nonnegative, which proves monotonicity.

\subsection{Proof of Proposition~\ref{prop:information-budget-curve}}

Monotonicity follows because every channel feasible at budget \(B_1\) is feasible at every \(B_2\geq B_1\). To prove convexity, choose two channels \(P_1(H_1\mid C,T)\) and \(P_2(H_2\mid C,T)\), and let \(Q\) be an independent Bernoulli time-sharing variable with \(\Pr(Q=1)=\lambda\). The combined message records \(Q\) and uses channel \(P_Q\), which gives
\begin{equation}
  I(C;Q,H_Q\mid T)
  =\lambda I(C;H_1\mid T)
  +(1-\lambda)I(C;H_2\mid T),
  \label{eq:information-budget-timeshare-rate}
\end{equation}
and
\begin{equation}
\begin{aligned}
  I(Y;C\mid Q,H_Q,X,T)
  ={}&\lambda I(Y;C\mid H_1,X,T)\\
  &+(1-\lambda)I(Y;C\mid H_2,X,T).
\end{aligned}
\label{eq:information-budget-timeshare-distortion}
\end{equation}
Applying the construction to channels arbitrarily close to the two infima proves convexity.

Nonnegativity of conditional mutual information and a constant message, which is feasible at every nonnegative budget and gives distortion \(I(Y;C\mid X,T)\), prove Equation~\eqref{eq:information-budget-basic-bounds}. At zero budget, \(I(C;H\mid T)=0\). Because the message is generated from \((C,T)\), \(H\) is then independent of \((C,X,Y)\) given \(T\), and conditioning on \(H\) does not change \(I(Y;C\mid X,T)\). It follows that \(D_{\mathrm{info}}(0)=I(Y;C\mid X,T)\). If \(C\) is discrete and \(B\geq H(C\mid T)\), the choice \(H=C\) is feasible and gives zero distortion.

For the final bound, exogeneity implies that \(X\) is independent of \((C,H)\) given \(T\), so
\begin{equation}
  I(C;H\mid X,T)=I(C;H\mid T).
  \label{eq:information-budget-exogenous-rate}
\end{equation}
The encoder Markov relation and the chain rule give
\begin{equation}
\begin{aligned}
  I(Y;C\mid X,T)
  ={}&I(Y;H\mid X,T)\\
  &+I(Y;C\mid H,X,T).
\end{aligned}
\label{eq:information-budget-chain}
\end{equation}
Data processing conditional on \((X,T)\) yields
\begin{equation}
  I(Y;H\mid X,T)
  \leq I(C;H\mid X,T)
  \leq B.
  \label{eq:information-budget-data-processing}
\end{equation}
Rearranging Equation~\eqref{eq:information-budget-chain} and using nonnegativity proves Equation~\eqref{eq:information-budget-linear-lower}.

\subsection{Proof of Proposition~\ref{prop:serialization-ordering}}

Let \(D_2\) be any decoder that uses \((H_2,X,T)\). Because a decoder that receives \(H_1\) can compute \(H_2=s(H_1,T)\) and then run \(D_2\), every risk achievable from \(H_2\) is also achievable from \(H_1\), which proves Equation~\eqref{eq:serialization-risk-ordering}. If \(H_1=r(H_2,T)\) also holds, the same argument in the opposite direction gives equality.

\subsection{Proof of Proposition~\ref{prop:unknown-query-cost}}

Because \(X\) is independent of \((C,H)\) and is uniform, we have
\begin{equation}
  I(Y;H\mid X)
  =\frac{1}{m}\sum_{j=1}^{m}I(C_j;H),
  \label{eq:unknown-query-average-information}
\end{equation}
where every information quantity in this proof is measured in bits. Independence of the coordinates gives
\begin{equation}
\begin{aligned}
  \sum_{j=1}^{m}I(C_j;H)
  &=m-\sum_{j=1}^{m}\mathsf H_2(C_j\mid H)\\
  &\leq m-\mathsf H_2(C\mid H)\\
  &=I(C;H)\\
  &\leq B.
\end{aligned}
\label{eq:unknown-query-information-budget}
\end{equation}
The inequality uses conditional subadditivity, \(\mathsf H_2(C\mid H)\leq\sum_j\mathsf H_2(C_j\mid H)\). Since \(Y\) is a fair bit conditional on \(X\), its conditional entropy satisfies
\begin{equation}
\begin{aligned}
  \mathsf H_2(Y\mid H,X)
  &=1-I(Y;H\mid X)\\
  &\geq1-\frac{B}{m}.
\end{aligned}
\end{equation}
Conditional entropy is nonnegative, which gives Equation~\eqref{eq:unknown-query-cost}. If the encoder observes \(X\), it sends \(C_X\), allowing the decoder to recover \(Y\) exactly from one bit.

\subsection{Proof of Proposition~\ref{prop:query-shift}}

Under the exogenous-query assumption, the conditional distribution of \((C,H)\) given \(T=t\) does not depend on the realized query. Define
\begin{equation}
  g_t(x)
  =\Ex\left[
    \KL\left(
      P_{Y\mid C,x,t}
      \mathbin\Vert
      P_{Y\mid H,x,t}
    \right)
    \mathrel{\big|}T=t
  \right].
  \label{eq:pointwise-query-information-loss}
\end{equation}
The function \(g_t\) is nonnegative. Conditioning on the query and applying Proposition~\ref{prop:log-loss-identity} gives
\begin{equation}
  \Delta_{\mu_t}(H)=\int g_t(x)\,d\mu_t(x),
  \qquad
  \Delta_{\nu_t}(H)=\int g_t(x)\,d\nu_t(x).
  \label{eq:query-shift-integrals}
\end{equation}
Absolute continuity and Equation~\eqref{eq:query-density-ratio} imply
\begin{equation}
\begin{aligned}
  \Delta_{\nu_t}(H)
  &=\int g_t(x)
    \frac{d\nu_t}{d\mu_t}(x)\,d\mu_t(x)\\
  &\leq\kappa\int g_t(x)\,d\mu_t(x)\\
  &=\kappa\Delta_{\mu_t}(H).
\end{aligned}
\end{equation}
Nonnegativity proves the remaining inequality in Equation~\eqref{eq:query-shift-bound}.

\subsection{Proof of Theorem~\ref{thm:bounded-loss-information}}
\label{app:proof-bounded-loss-information}

For a posterior distribution \(p\) on \(Y\), define its Bayes envelope by
\begin{equation}
  V(p)=\min_{a\in\mathcal A}\Ex_p[\ell(a,Y)].
  \label{eq:bayes-envelope}
\end{equation}
If \(p\) and \(q\) are two distributions, let \(a_q\) minimize the risk under \(q\). Since \(0\leq\ell\leq L_{\max}\), the Bayes-envelope difference satisfies
\begin{equation}
\begin{aligned}
  V(p)-V(q)
  &\leq \Ex_p[\ell(a_q,Y)]-
  \Ex_q[\ell(a_q,Y)]\\
  &\leq L_{\max}\TV(p,q).
\end{aligned}
\label{eq:bayes-envelope-tv}
\end{equation}
Interchanging \(p\) and \(q\) gives
\begin{equation}
  |V(p)-V(q)|\leq L_{\max}\TV(p,q).
  \label{eq:bayes-envelope-lipschitz}
\end{equation}

Let
\begin{equation}
  p_C=P_{Y\mid C,X,T},
  \qquad
  p_H=P_{Y\mid H,X,T}.
  \label{eq:posterior-c-h}
\end{equation}
The encoder Markov condition gives
\begin{equation}
  p_H=\Ex[p_C\mid H,X,T].
  \label{eq:posterior-mixture}
\end{equation}
Because \(V\) is the pointwise minimum of linear functions of \(p\), it is concave. Conditional Jensen's inequality applied to Equation~\eqref{eq:posterior-mixture} proves \(\RE\geq\Rfull\).

Equation~\eqref{eq:bayes-envelope-lipschitz} also yields
\begin{equation}
  \RE-\Rfull
  \leq L_{\max}\Ex[\TV(p_C,p_H)].
  \label{eq:risk-tv}
\end{equation}
Pinsker's inequality and Jensen's inequality for the square root give
\begin{equation}
\begin{aligned}
  \Ex[\TV(p_C,p_H)]
  &\leq
  \Ex\left[
    \sqrt{\frac{1}{2}\KL(p_C\Vert p_H)}
  \right]\\
  &\leq
  \sqrt{\frac{1}{2}
  \Ex[\KL(p_C\Vert p_H)]}\\
  &=\sqrt{\frac{I(Y;C\mid H,X,T)}{2}}.
\end{aligned}
\label{eq:pinsker-information}
\end{equation}
Combining Equations~\eqref{eq:risk-tv} and~\eqref{eq:pinsker-information} proves Equation~\eqref{eq:bounded-loss-information}.

\subsection{Proof of Proposition~\ref{prop:mixed-state-description}}

Quantize each coordinate of \(\theta\) on \([0,1]\) with a grid whose reconstruction error is at most
\begin{equation}
  q=\frac{\varepsilon}{L_S\sqrt d}.
  \label{eq:mixed-state-grid}
\end{equation}
The grid needs at most \(1+1/q\) codewords per coordinate, and the reconstructed vector \(\overline\theta\) satisfies
\begin{equation}
  \lVert\overline\theta-\theta\rVert_2
  \leq\sqrt d q
  =\frac{\varepsilon}{L_S}.
\end{equation}
Encode every coordinate of \(m\) exactly. The Lipschitz assumption then bounds the risk increase by \(\varepsilon\), and summing the fixed-length coordinate and discrete-field codes gives Equation~\eqref{eq:mixed-state-description}.

\subsection{Proof of Theorem~\ref{thm:state-packing}}

For each realized message \(h\), let \(\pi_h\) denote the conditional action distribution induced by the decoder after integrating over its private randomization. Thus, \(\pi_h\) maps the future observation \(X\) to an action distribution. Let \(\Pi=\pi_H\) be the random policy selected by the boundary message. The policy is fixed before the realized \(X\) is observed, although its action distribution may depend on \(X\) when the policy is executed. Because \(\Pi\) is a deterministic function of \(H\), data processing gives
\begin{equation}
  I(J;\Pi\mid T=t)
  \leq I(J;H\mid T=t)
  \leq B.
  \label{eq:policy-data-processing}
\end{equation}
Define \(\widehat J(\Pi)\) by returning the unique index for which \(\Pi\) is \(\delta\)-near-optimal and returning an arbitrary index otherwise. Because the packing condition guarantees uniqueness, every event \(\widehat J\neq J\) satisfies
\begin{equation}
  L_J(\Pi)-L_J^\star\geq\delta.
  \label{eq:packing-error-regret}
\end{equation}
Therefore, we have
\begin{equation}
  \Ex[\Regret]
  \geq\delta\Pr(\widehat J\neq J).
  \label{eq:packing-regret-testing}
\end{equation}
Fano's inequality and Equation~\eqref{eq:policy-data-processing} imply
\begin{equation}
  \Pr(\widehat J\neq J)
  \geq
  1-\frac{B+\log 2}{\log N}.
  \label{eq:packing-fano}
\end{equation}
A probability is nonnegative, so combining Equations~\eqref{eq:packing-regret-testing} and~\eqref{eq:packing-fano} proves Equation~\eqref{eq:state-packing-lower}.

\section{Proofs for the Parametric Results}
\label{app:parametric-proofs}

This appendix proves the results for linear regression with Gaussian noise. The proofs first establish exact sufficiency and then treat the alternative presentation and finite-precision variants.

\subsection{Proof of Theorem~\ref{thm:linear-sufficiency}}

The Gaussian likelihood can be written as
\begin{equation}
\begin{aligned}
  p(y_n\mid X_n,\beta)
  &\propto
  \exp\left(
    -\frac{1}{2\sigma^2}
    \lVert y_n-X_n\beta\rVert_2^2
  \right)\\
  &\propto
  \exp\left(
    -\frac{1}{2\sigma^2}
    \left(
      \beta^{\top}G_n\beta-2\beta^{\top}b_n
    \right)
  \right).
\end{aligned}
\label{eq:linear-likelihood-statistics}
\end{equation}
The omitted factor depends on \(y_n^{\top}y_n\) but not on \(\beta\). Multiplying Equation~\eqref{eq:linear-likelihood-statistics} by the Gaussian prior and completing the square gives Equation~\eqref{eq:linear-posterior}. The conditional distribution of \(\beta\) given \((X_n,y_n)\) therefore depends on the demonstrations only through \((G_n,b_n)\).

For a future query \(x\), integrating \(\mathcal N(x^{\top}\beta,\sigma^2)\) against that posterior gives the conditional distribution of \(Y\) in Equation~\eqref{eq:linear-predictive}, which is determined by \(H_n\). It follows that
\begin{equation}
  P(Y\mid X_n,y_n,H_n,x,T)
  =P(Y\mid H_n,x,T),
\end{equation}
which proves predictive sufficiency and Equation~\eqref{eq:linear-mutual-information}.

\subsection{Proof of Theorem~\ref{thm:synthetic-sufficient-equivalence}}

Every real matrix \(X_n\) satisfies
\begin{equation}
  \operatorname{range}(X_n^{\top}X_n)
  =\operatorname{range}(X_n^{\top}).
  \label{eq:gram-range}
\end{equation}
Since \(b_n=X_n^{\top}y_n\), Equation~\eqref{eq:gram-range} implies \(b_n\in\operatorname{range}(G_n)\). Therefore, we have
\begin{equation}
  U_rU_r^{\top}b_n=b_n.
  \label{eq:b-in-gram-range}
\end{equation}
Using Equation~\eqref{eq:synthetic-sufficient-construction}, we obtain
\begin{equation}
\begin{aligned}
  \widetilde X^{\top}\widetilde X
  &=U_r\Lambda_r^{1/2}
    \Lambda_r^{1/2}U_r^{\top}
  =G_n,\\
  \widetilde X^{\top}\widetilde y
  &=U_r\Lambda_r^{1/2}
    \Lambda_r^{-1/2}U_r^{\top}b_n
  =b_n.
\end{aligned}
\end{equation}
Equation~\eqref{eq:synthetic-sufficient-statistics} follows. Because Equation~\eqref{eq:linear-posterior} depends on the data only through these two quantities, the Gaussian posterior and posterior predictive distributions agree, and substituting Equation~\eqref{eq:synthetic-sufficient-statistics} into the two ridge estimators proves Equation~\eqref{eq:ridge-synthetic-sufficient-equivalence}.

\subsection{Proof of Corollary~\ref{cor:synthetic-sufficient-rank-minimal}}

Every real matrix \(Z\) satisfies
\begin{equation}
  \rank(Z^{\top}Z)=\rank(Z)\leq s.
\end{equation}
If \(Z^{\top}Z=G_n\), then \(s\geq\rank(G_n)\). Equation~\eqref{eq:synthetic-sufficient-construction} uses exactly \(r=\rank(G_n)\) rows, so it attains this lower bound.

\subsection{Proof of Proposition~\ref{prop:truncated-ridge}}

Let \(P_\tau=U_\tau U_\tau^{\top}\). A singular-value decomposition of \(X_n\) shows that
\begin{equation}
  \lVert(I-P_\tau)b_n\rVert_2
  \leq \sqrt\tau\lVert y_n\rVert_2.
  \label{eq:truncated-b-bound}
\end{equation}
If \(X_n=Q\Sigma U^{\top}\), then \(b_n=U\Sigma Q^{\top}y_n\), and every discarded singular value is at most \(\sqrt\tau\).

The matrices \(G_n\), \(G_{n,\tau}\), and \(P_\tau\) share the same eigenvectors. The two ridge estimators have identical coefficients on the retained eigenspace, but the truncated estimator is zero on the discarded eigenspace. Their difference is therefore
\begin{equation}
  \widehat\beta_\lambda-
  \widehat\beta_{\lambda,\tau}
  =(G_n+\lambda I)^{-1}(I-P_\tau)b_n.
  \label{eq:truncated-ridge-decomposition}
\end{equation}
Because the inverse norm is at most \(1/\lambda\), taking norms and applying Equation~\eqref{eq:truncated-b-bound} proves Equation~\eqref{eq:truncated-ridge-bound}.

\subsection{Proof of Theorem~\ref{thm:linear-quantization}}

Because \(G_n\succeq0\) and \(\overline G\succeq0\), the matrices \(A\) and \(\overline A\) satisfy
\begin{equation}
  A\succeq\alpha I,
  \qquad
  \overline A\succeq\alpha I.
  \label{eq:posterior-precision-lower}
\end{equation}
Hence, \(\lVert V\rVert_{\mathrm{op}}\leq1/\alpha\) and \(\lVert\overline V\rVert_{\mathrm{op}}\leq1/\alpha\). The resolvent identity gives
\begin{equation}
  \overline V-V
  =\overline V(A-\overline A)V.
  \label{eq:posterior-resolvent}
\end{equation}
Since
\begin{equation}
  \lVert A-\overline A\rVert_{\mathrm{op}}
  =\sigma^{-2}
  \lVert G_n-\overline G\rVert_{\mathrm{op}},
\end{equation}
Equation~\eqref{eq:covariance-quantization-bound} follows.

For the posterior mean, we have
\begin{equation}
\begin{aligned}
  \overline m-m
  &=\overline V(\overline h-h)
  +(\overline V-V)h.
\end{aligned}
\label{eq:mean-perturbation}
\end{equation}
The first term is at most \(\delta_b/(\sigma^2\alpha)\), and the second is at most \(\delta_G\lVert h\rVert_2/(\sigma^2\alpha^2)\), which establishes Equation~\eqref{eq:mean-quantization-bound}. The two inequalities in Equation~\eqref{eq:predictive-quantization-bound} follow from Cauchy--Schwarz and the definition of the operator norm.

\subsection{Proof of Corollary~\ref{cor:linear-predictive-kl}}

The predictive variances satisfy \(v_x\geq\sigma^2\) and \(\overline v_x\geq\sigma^2\) because both posterior covariance matrices are positive semidefinite. Theorem~\ref{thm:linear-quantization} gives
\begin{equation}
  |\mu_x-\overline\mu_x|\leq\varepsilon_{\mu},
  \qquad
  |v_x-\overline v_x|\leq\varepsilon_v.
  \label{eq:predictive-parameter-errors}
\end{equation}
For the divergence in the direction used in Corollary~\ref{cor:linear-predictive-kl}, the univariate Gaussian formula is
\begin{equation}
\begin{aligned}
  &\KL\!\left(
    \mathcal N(\mu_x,v_x)
    \mathbin\Vert
    \mathcal N(\overline\mu_x,\overline v_x)
  \right)\\
  &\quad=\frac{1}{2}\left(
    \log\frac{\overline v_x}{v_x}
    +\frac{v_x}{\overline v_x}-1
    +\frac{(\mu_x-\overline\mu_x)^2}{\overline v_x}
  \right).
  \label{eq:gaussian-predictive-kl}
\end{aligned}
\end{equation}
Let \(t=v_x/\overline v_x\). Since \(\overline v_x\geq\sigma^2\), Equation~\eqref{eq:predictive-parameter-errors} gives \(|t-1|=|v_x-\overline v_x|/\overline v_x\leq\varepsilon_v/\sigma^2\leq1/2\). On \([1/2,3/2]\), the function \(t-1-\log t\) has second derivative at most four and vanishes together with its first derivative at one, so
\begin{equation}
  t-1-\log t\leq2(t-1)^2
  \leq\frac{2\varepsilon_v^2}{\sigma^4}.
  \label{eq:variance-kl-bound}
\end{equation}
The mean term in Equation~\eqref{eq:gaussian-predictive-kl} is at most \(\varepsilon_{\mu}^2/\sigma^2\). Multiplying the two bounds by one half proves Equation~\eqref{eq:predictive-kl-quantization}.

\subsection{Proof of Corollary~\ref{cor:linear-bit-budget}}

Every entry of \(G_n\) lies in \([-nL^2,nL^2]\), and every entry of \(b_n\) lies in \([-nLB_y,nLB_y]\). A uniform scalar quantizer with step \(q\) on an interval of width \(2A\) needs at most
\begin{equation}
  \left\lceil\log_2\left(1+\frac{2A}{q}\right)\right\rceil
  \label{eq:scalar-quantizer-count}
\end{equation}
bits, apart from a fixed encoding header. Storing the upper triangle of the symmetric matrix and the \(d\) entries of \(b_n\) gives Equation~\eqref{eq:linear-bit-budget}.

Before projection, every reconstructed matrix entry differs from the corresponding entry of \(G_n\) by at most \(q_G\), so the Frobenius norm of the full symmetric error is at most \(q_Gd\) and the operator norm is no larger. Projection onto the positive semidefinite cone is a metric projection in Frobenius norm and therefore cannot increase the distance to the positive semidefinite matrix \(G_n\). Together with the vector bound \(q_b\sqrt d\), these observations prove Equation~\eqref{eq:linear-entrywise-errors}.

\subsection{Proof of Corollary~\ref{cor:linear-bit-kl-rate}}

Let \(m=\lfloor(B-B_0)/p\rfloor\), where \(B_0\) accounts for the fixed header, signs, dimensions, and coding conventions, and quantize every stored scalar using \(m\) bits over its bounded interval. Corollary~\ref{cor:linear-bit-budget} then gives constants \(c_G\) and \(c_b\), depending only on the fixed problem parameters, such that
\begin{equation}
  \delta_G\leq c_G2^{-m},
  \qquad
  \delta_b\leq c_b2^{-m}.
  \label{eq:bit-rate-parameter-errors}
\end{equation}
Equations~\eqref{eq:predictive-kl-errors} and~\eqref{eq:predictive-kl-quantization} therefore give a constant \(K_0\) for which the predictive divergence is at most \(K_0 2^{-2m}\) once \(B_0\) is large enough to ensure \(\varepsilon_v\leq\sigma^2/2\). Since \(m\geq(B-B_0)/p-1\), increasing the constant by a factor of four proves Equation~\eqref{eq:linear-bit-kl-rate}.

\section{Proofs for the Nonparametric Results}
\label{app:nonparametric-proofs}

This appendix proves the upper and lower bounds for the nonparametric regression setting. The argument first controls the cell-based record and then constructs a packing for the memory lower bound.

\subsection{Proof of Theorem~\ref{thm:nonparametric-upper}}

Let \(h=M^{-1/d}\) be the side length of each cell, and let
\begin{equation}
  D_M=\sqrt d\,h
  \label{eq:cell-diameter}
\end{equation}
be its diameter. Write \(p_j=\Pr(X\in A_j)\) and
\begin{equation}
  f_j=\Ex[f(X)\mid X\in A_j].
  \label{eq:cell-population-mean}
\end{equation}
The density lower bound gives \(p_j\geq p_{\min}/M\).

When \(N_j>0\), the cell predictor is \(Q_q(\overline Y_j)\). For \(x\in A_j\), the H\"older condition gives
\begin{equation}
  |f_j-f(x)|\leq LD_M^\beta.
  \label{eq:cell-bias}
\end{equation}
The conditional variance of \(Y\) in a cell is at most \(B_y^2\). Conditional on \(N_j=k>0\), the sampling error therefore satisfies
\begin{equation}
  \Ex[(\overline Y_j-f_j)^2\mid N_j=k]
  \leq\frac{B_y^2}{k}.
  \label{eq:cell-mean-variance}
\end{equation}

For a binomial variable \(N\sim\operatorname{Binomial}(n,p)\), we have \(1/N\leq2/(N+1)\) on \(N>0\), and
\begin{equation}
  \Ex\left[\frac{1}{N+1}\right]
  =\frac{1-(1-p)^{n+1}}{(n+1)p}
  \leq\frac{1}{(n+1)p}.
  \label{eq:binomial-inverse}
\end{equation}
It follows that
\begin{equation}
  \Ex\left[\frac{\mathbf 1(N_j>0)}{N_j}\right]
  \leq\frac{2}{(n+1)p_j}.
  \label{eq:binomial-positive-inverse}
\end{equation}

On a nonempty cell, use
\begin{equation}
\begin{aligned}
  Q_q(\overline Y_j)-f(x)
  ={}&
  (Q_q(\overline Y_j)-\overline Y_j)\\
  &+(\overline Y_j-f_j)+(f_j-f(x)).
\end{aligned}
\label{eq:cell-error-decomposition}
\end{equation}
The inequality \((a+b+c)^2\leq3(a^2+b^2+c^2)\), Equations~\eqref{eq:cell-bias} and~\eqref{eq:cell-mean-variance}, and the quantizer bound imply
\begin{equation}
\begin{aligned}
  &\Ex\left[
    (Q_q(\overline Y_j)-f(x))^2
    \mathbf 1(N_j>0)
  \right]\\
  &\qquad\leq
  3q^2+3L^2D_M^{2\beta}
  +3B_y^2
  \Ex\left[
    \frac{\mathbf 1(N_j>0)}{N_j}
  \right].
\end{aligned}
\label{eq:nonempty-cell-risk}
\end{equation}
Integrating Equation~\eqref{eq:nonempty-cell-risk} over \(A_j\) and summing over cells gives a variance contribution at most \(6B_y^2M/(n+1)\), together with the bias and quantization terms.

On an empty cell, the predictor is zero and the squared error is at most \(B_f^2\). Since
\begin{equation}
  \Pr(N_j=0)=(1-p_j)^n
  \leq\exp(-np_j),
  \label{eq:empty-cell-probability}
\end{equation}
the integrated empty-cell contribution is at most
\begin{equation}
\begin{aligned}
  \sum_{j=1}^{M}
  p_jB_f^2\exp(-np_j)
  &\leq
  B_f^2\exp\left(-\frac{np_{\min}}{M}\right).
\end{aligned}
\label{eq:empty-cell-risk}
\end{equation}
Finally, \(D_M^{2\beta}=d^\beta M^{-2\beta/d}\). Combining the nonempty and empty contributions proves Equation~\eqref{eq:nonparametric-upper} with explicit constants obtained from the preceding bounds.

\subsection{Proof of Corollary~\ref{cor:nonparametric-bit-budget}}
\label{app:proof-nonparametric-bit-budget}

A count in \(\{0,1,\ldots,n\}\) uses \(\lceil\log_2(n+1)\rceil\) bits, and a quantized mean in \([-B_y,B_y]\) with step \(q\) has at most \(1+2B_y/q\) levels. The total number of bits is therefore at most
\begin{equation}
  M\left(
    \left\lceil\log_2(n+1)\right\rceil
    +
    \left\lceil
      \log_2\left(1+\frac{2B_y}{q}\right)
    \right\rceil
    +c_0
  \right),
  \label{eq:cell-total-bits}
\end{equation}
where \(c_0\) accounts for fixed field delimiters and the empty-cell flag. Under Equation~\eqref{eq:nonparametric-optimal-choice}, both logarithmic terms are \(O(\log n)\), and \(M=O(n^{d/(2\beta+d)})\). These bounds establish Equation~\eqref{eq:nonparametric-bit-budget}. Substitution into Theorem~\ref{thm:nonparametric-upper} gives the risk order in Corollary~\ref{cor:nonparametric-bit-budget}.

\subsection{Proof of Corollary~\ref{cor:nonparametric-budget-upper}}

Choosing \(q^2\) proportional to \(M^{-2\beta/d}\) ensures, because \(M\leq n\), that each count and quantized mean uses \(O(\log(n+1))\) bits. A sufficiently small constant \(c_{\mathrm{code}}\) then makes every admissible \(M=m^d\leq M_B\) feasible, and substituting into Equation~\eqref{eq:nonparametric-upper} before minimizing over the feasible partitions proves Equation~\eqref{eq:nonparametric-budget-upper}.

\subsection{A Packing Lemma for H\"older Functions}

A standard bump construction underlying Assouad and Fano lower bounds \citep{Tsybakov2008introductionnonparametric} supplies constants \(a_0,c_0,c_1>0\) such that, for every integer \(m\), setting \(M=m^d\) and \(h=1/m\) yields a family
\begin{equation}
  \{f_\theta:\theta\in\{-1,1\}^{M}\}
  \subseteq\mathcal H^\beta(L,B_f)
  \label{eq:holder-hypercube}
\end{equation}
such that, whenever \(\theta\) and \(\theta'\) differ in one coordinate,
\begin{equation}
  \lVert f_\theta-f_{\theta'}\rVert_2^2
  =c_0h^{2\beta+d},
  \label{eq:adjacent-holder-distance}
\end{equation}
and, for arbitrary \(\theta,\theta'\),
\begin{equation}
  \lVert f_\theta-f_{\theta'}\rVert_2^2
  =c_0h^{2\beta+d}
  d_{\mathrm H}(\theta,\theta'),
  \label{eq:holder-hamming-distance}
\end{equation}
where \(d_{\mathrm H}\) is Hamming distance. Because the construction uses disjoint, scaled H\"older bumps with amplitude \(a_0h^\beta\), choosing \(a_0\) sufficiently small keeps the entire family inside the H\"older ball.

\subsection{Proof of Theorem~\ref{thm:nonparametric-lower}}

We first prove the sample floor. Under the binary-response submodel in Equation~\eqref{eq:bounded-binary-submodel}, let
\begin{equation}
  p_f(x)=\frac{1}{2}\left(1+\frac{f(x)}{B_y}\right).
\end{equation}
The packing construction keeps \(p_f(x)\in[1/4,3/4]\). The Bernoulli divergence bound
\begin{equation}
  \KL(\operatorname{Ber}(p)\Vert\operatorname{Ber}(q))
  \leq\frac{(p-q)^2}{q(1-q)}
  \label{eq:bernoulli-kl-bound}
\end{equation}
therefore implies that, for a constant \(c_B\) depending only on \(B_y\),
\begin{equation}
  \KL(P_f^{(n)}\Vert P_g^{(n)})
  \leq c_Bn\lVert f-g\rVert_2^2.
  \label{eq:bounded-regression-kl}
\end{equation}
For adjacent vertices in the family of Equation~\eqref{eq:holder-hypercube}, Equations~\eqref{eq:adjacent-holder-distance} and~\eqref{eq:bounded-regression-kl} give divergence of order \(nh^{2\beta+d}\). Choosing \(h\) proportional to \(n^{-1/(2\beta+d)}\) with a sufficiently small constant bounds the adjacent divergence by a fixed number smaller than one, so Assouad's lemma yields
\begin{equation}
  \inf_{\widehat f}
  \sup_{f\in\mathcal H^\beta(L,B_f)}
  \Ex_f[\lVert\widehat f-f\rVert_2^2]
  \geq c_2h^{2\beta}
  \geq c_3n^{-2\beta/(2\beta+d)}.
  \label{eq:sample-floor-proof}
\end{equation}
This lower bound applies to every handover because it also applies when the decoder receives the full data.

We next prove the memory floor. The Varshamov--Gilbert bound supplies a subset \(\Theta\subseteq\{-1,1\}^{M}\) such that
\begin{equation}
  \log|\Theta|\geq c_4M
  \label{eq:vg-cardinality}
\end{equation}
and every distinct pair in \(\Theta\) has Hamming distance at least \(M/8\). Equation~\eqref{eq:holder-hamming-distance} then gives pairwise squared separation at least
\begin{equation}
  c_5h^{2\beta+d}M
  =c_5h^{2\beta}.
  \label{eq:packing-separation}
\end{equation}
Let \(J\) be uniform on \(\Theta\). Whatever data the encoder observes, a message in an alphabet of size at most \(2^B\) satisfies
\begin{equation}
  I(J;H)\leq B\log 2.
  \label{eq:message-information-bound}
\end{equation}
Any randomized decoder can be represented as \(D(H,U,x)\), where the private random seed \(U\) is independent of \((J,H)\). Conditional on \((H,U)\), the map \(x\mapsto D(H,U,x)\) defines a reconstructed function, and
\begin{equation}
  I(J;H,U)=I(J;H)\leq B\log 2.
  \label{eq:decoder-randomness-information}
\end{equation}
We decode \(J\) by the closest element of the packing in \(L_2\). The pairwise separation in Equation~\eqref{eq:packing-separation} implies that an incorrect nearest-packing decision requires squared estimation error of at least one quarter of that separation. Fano's inequality then gives a constant \(c_6>0\) such that
\begin{equation}
  \inf_{E,D}
  \sup_{f\in\mathcal H^\beta(L,B_f)}
  \Ex_f[\lVert\widehat f_{E,D}-f\rVert_2^2]
  \geq c_6h^{2\beta}
  \left(
    1-\frac{B\log 2+\log 2}{c_4M}
  \right).
  \label{eq:fano-memory-bound}
\end{equation}
We choose \(m\) so that \(M=m^d\) is the smallest admissible order of \(B+1\) for which the expression in parentheses is bounded below by a positive constant. Since \(h=M^{-1/d}\), Equation~\eqref{eq:fano-memory-bound} becomes
\begin{equation}
  R_{n,B}\geq c_7(B+1)^{-2\beta/d}.
  \label{eq:memory-floor-proof}
\end{equation}
The risk is at least both Equations~\eqref{eq:sample-floor-proof} and~\eqref{eq:memory-floor-proof}. Reducing the constant proves Equation~\eqref{eq:nonparametric-lower}.